\documentclass[nohyperref]{article}

\usepackage{microtype}
\usepackage{graphicx}
\usepackage{subfigure}
\usepackage{booktabs} 
\usepackage{soul}

\usepackage{hyperref}

\usepackage[accepted]{icml2023}

\usepackage{amsmath}
\usepackage{amssymb}
\usepackage{mathtools}
\usepackage{amsthm}

\usepackage[capitalize,noabbrev]{cleveref}

\theoremstyle{plain}

\theoremstyle{definition}

\theoremstyle{remark}

\usepackage[textsize=tiny]{todonotes}

\usepackage{graphicx}
\usepackage{float}
\usepackage{xcolor}
\usepackage{wrapfig}
\usepackage{graphicx}

\def \ubf{{\mathbf u}}
\def \z{{\mathbf z}}
\def \w{{\mathbf w}}
\def \bomega{{\boldsymbol\omega}}
\def \bX{{\mathbf X}}
\def \bsigma{{\boldsymbol\sigma}}
\def \cF {{\mathcal{F}}}
\def \bV{{\mathbf{V}}}
\def \r{{\mathbf r}}
\def \md{{\mathrm d}}
\def \NPDF{{\mathcal N}}

\def \cG{{\mathcal{G}}}
\def \cB{{\mathcal{B}}}
\def \Rb{{\mathbb R}}
\def \Eb{{\mathbb E}}
\def \bSigma{{\boldsymbol \Sigma}}
\def \blambda{{\boldsymbol\lambda}}
\def \bLambda{{\boldsymbol\Lambda}}
\def \bV{{\mathbf V}}
\def \bepsilon{{\boldsymbol \epsilon}}
\def \bmu{{\boldsymbol \mu}}
\def \ba{{\mathbf a}}
\def \bI{{\mathbf I}}
\def \b0{{\mathbf 0}}

\def \y{{\mathbf y}}

\def \cGP{{\mathcal{GP}}}

\def \b0{{\mathbf{0}}}

\def \bmu {{\boldsymbol \mu}}

\def \cU {{\mathcal U}}
\def \ebf{{\mathbf e}}
\def \bK {{\mathbf K}}

\def \obf {{\mathbf o}}
\def \Obf {{\mathbf O}}
\def \cO {{\mathcal O}}

\icmltitlerunning{Random Grid Neural Processes for Parametric Partial Differential Equations}

\begin{document}

\twocolumn[


\icmltitle{Random Grid Neural Processes for Parametric Partial Differential Equations}



\icmlsetsymbol{equal}{*}

\begin{icmlauthorlist}
\icmlauthor{Arnaud Vadeboncoeur}{1}
\icmlauthor{Ieva Kazlauskaite}{1}
\icmlauthor{Yanni Papandreou}{2}
\icmlauthor{Fehmi Cirak}{1}
\icmlauthor{Mark Girolami}{1,3}
\icmlauthor{\"Omer Deniz Akyildiz}{2}
\end{icmlauthorlist}

\icmlaffiliation{1}{Department of Engineering, University of Cambridge, Trumpington St, Cambridge CB2 1PZ.}
\icmlaffiliation{2}{Department of Mathematics, Imperial College London, Exhibition Rd, South Kensington, London SW7 2AZ, United Kingdom.}
\icmlaffiliation{3}{The Alan Turing Institute, British Library, 96 Euston Rd, London NW1 2DB, United Kingdom.}

\icmlcorrespondingauthor{Arnaud Vadeboncoeur}{av537@cam.ac.uk}

\icmlkeywords{Physics Informed, Variational Inference, PDE, Neural Process, Gaussian Process, Inverse Problems}

\vskip 0.3in
]



\printAffiliationsAndNotice{}  

\begin{abstract}
We introduce a new class of spatially stochastic physics and data informed deep latent models for parametric partial differential equations (PDEs) which operate through scalable variational neural processes. We achieve this by assigning probability measures to the spatial domain, which allows us to treat collocation grids probabilistically as random variables to be marginalised out. Adapting this spatial statistics view, we solve forward and inverse problems for parametric PDEs in a way that leads to the construction of Gaussian process models of solution fields. The implementation of these random grids poses a unique set of challenges for inverse physics informed deep learning frameworks and we propose a new architecture called Grid Invariant Convolutional Networks (GICNets) to overcome these challenges. We further show how to incorporate noisy data in a principled manner into our physics informed model to improve predictions for problems where data may be available but whose measurement location does not coincide with any fixed mesh or grid. The proposed method is tested on a nonlinear Poisson problem,  Burgers equation, and Navier-Stokes equations, and we provide extensive numerical comparisons. We demonstrate significant computational advantages over current physics informed neural learning methods for parametric PDEs while improving the predictive capabilities and flexibility of these models.
\end{abstract}





\section{Introduction}
Partial differential equations (PDEs) are of central importance in natural sciences and engineering as models for describing physical phenomena. Numerical solvers of these equations have received immense attention over the decades with the development of methods such as finite differences, finite elements, spectral methods, to name a few \cite{quarteroni2008numerical}. 
The main computational problems pertinent to PDE modelling can be categorised into two main classes: The task of obtaining a solution to a PDE for a given set of parameters (\textit{forward problem}) and the task of recovering model parameters from solution fields or observations of solution fields (\textit{inverse problem}) \cite{belov2012inverse, stuart_2010}.
Solving each of these problems efficiently and accurately remains an open problem in the field of scientific computing. Both problems are of fundamental importance in scientific and engineering applications and creating efficient and scalable methods for solving these problems could have large ramifications for engineering practice.
Recently, physics informed machine learning (ML) \cite{raissi2019physics} offers a novel perspective on solving these types of problems for parametric PDEs and has been shown to be advantageous to train large networks to learn parametric PDEs for ranges of parameters \cite{lu2021learning, li2020fourier}. These models can then be deployed to solve PDEs in real-time. 
The central objective of this paper is to introduce accurate and uncertainty aware physics informed probabilistic deep learning methods that go beyond \textit{fixed grid} approaches to parametric PDEs, allowing for a synthesis with \textit{noisy} data measured on arbitrary grids for forward and inverse problems. 

\subsection{Contributions}
In this paper, we propose a new framework for jointly learning probabilistic mappings of forward and inverse problems of parametric PDEs using ideas from spatial statistics \cite{ripley2005spatial, cressie2021spatial}. This new approach connects neural Gaussian process models of PDE solution fields with physical parameters through conditioning on random domain partitions. More precisely, we propose (1) a physics driven variational inference framework based on random grids; (2) new kernels for the learning of Gaussian random fields; (3) a new grid invariant architecture to enable learning through random collocation. We demonstrate our methods on three PDEs of increasing complexity. The first is a 1D nonlinear Poisson PDE where we learn the mapping of a spectral representation of a diffusion field to the solution field for a range of forcing conditions. The second PDE is the spatio-temporal Burgers equation learned for a range of diffusion and non-linearity coefficients and of parameterized initial conditions. The third PDE is the incompressible Navier-Stokes lid-driven cavity flow problem where we map density and viscosity coefficients to spatio-temporal 3D solution fields. Furthermore, we demonstrate how to correctly incorporate sparse, \textit{noisy} observations of sample solution fields to improve the predictive capabilities of the model. We furthermore compare our proposed method to existing methods such as Physics Driven Deep Latent Variable Models (PDDLVM) \cite{vadeboncoeur2022deep}, modified DeepONets \cite{lu2021learning}, and Physics-Informed Parametric Fourier Feature Networks \cite{wang2021eigenvector} which we adapt to solve both forward and inverse problems as described in \cite{zhao2022learning}.

\subsection{Related Work}
In this section, we review relevant work in the literature. We focus our review on machine learning methods for inference in parametric PDEs.


\textbf{Supervised and Semi-Supervised Operator Learning:}
These methods for learning differential operators are based either entirely or partially on \textit{noiseless} PDE solution datasets created with classical numerical methods such as FEM or spectral methods. Such methods include Fourier Neural Operators \cite{li2020fourier, fanaskov2022spectral, tripura2023wavelet}. Other methods which use a combination of pre-computed solutions from classical solvers and physics informed losses include Physics Informed Neural Operators and DeepONets \cite{li2021physics, lu2021learning}. Although these methods have been shown to be effective in certain scenarios, they fundamentally rely on classical solvers in order to learn, and so are bound to their natural limitations for learning new problems. The limitations include the need to recompute models from scratch for new parameter instances and rely on CPU-driven operations that do not parallelize easily. Furthermore, these methods generally focus on learning the operator from one function space to another and are typically evaluated on fixed grids. This is restrictive in cases where the domain geometry, boundary, and initial conditions cannot be easily defined in a function space form. Instead, we focus our work on PDEs parameterized by sets of scalar coefficients. 
The parametric representation overcomes many of the limitations outlined in \citet{lu2022comprehensive, tang2023physics}. 

\textbf{Physics Informed Variational Autoencoders:}
Several methods have been proposed to adapt the popular Variational Autoencoder (VAE) \cite{welling2014auto} framework to physics problems. Such methods include physics-informed VAEs \cite{zhong2022pi}, physics integrated VAEs \cite{takeishi2021physics}, autoencoding  
 PDEs \cite{tait2020variational}, and physics-informed dynamical VAEs \cite{glyn2022phi}. While these methods rely on variational inference like our approach, they aim at solving a fundamentally different problem as they relate the observable space to a \textit{discovered} physical latent space. In contrast, we relate solution fields to physical latent spaces and may use some data from the observation space to enhance these mappings but we do not directly map parameters to the observation space (but rather to the solution space). Furthermore, we are interested in methods that can perform inference in the absence of data and are only supplemented/improved by data.

\textbf{Neural Processes:}
Neural processes are a new class of deep probabilistic regression tools \cite{garnelo2018neural}. Neural and conditional neural processes \cite{markou2022practical, garnelo2018conditional} output Gaussian uncertainty estimates around predictions similar to Gaussian processes, but they do not suffer from the same training scalability issues. These have partially been adapted to physics problems in \citet{yang2019conditional, yang2019adversarial}, however, they are essentially data-driven rather than physics driven methods. 

Methods related to the probabilistic modelling of single instances of solution fields are pertinent to the proposed methodology, such as \cite{pmlr-v162-long22a, pmlr-v162-tronarp22a}. Further methods such as \cite{lu2021physics} compute inverse problems by posing an optimisation problem. Other works parameterize physics informed Gaussian processes \cite{pang2019neural, pmlr-v162-long22a, zhang2022pagp, chen2021solving}. These methods are not adapted to parametric PDE scenarios as described in \citet{bhattacharya2020model}. We also note the work of \citet{ardizzone2018analyzing} which uses invertible networks to solve inverse problems and the methods used for discovering dynamics, e.g., \citet{raissi2018hidden, brunton2016discovering}. However, these methods have different objectives than solving classic inverse problems and focus on hidden dynamics and discovering coordinate transformations. Other works \cite{chiu2022can} have explored the use of sets of random collocation points drawn for every residual evaluation, but this was studied for single solution field instances, not parametric PDEs, and only for the forward problem.
%


\section{Background}\label{sec:background}
\subsection{Physics Informed ML for Forward Problems}
We formulate the nonlinear parametric PDE of interest as 
\begin{align}
    \cG^\w_\z(u)(x) &= 0, \quad x\in\Omega, \label{eq:PDE} \\
    \cB_\w(u)(x) &= 0, \quad x\in \partial\Omega, \label{eq:BoundaryCond}
\end{align}
where $\Omega \subset \Rb^d$, $\cG^\w_\z$ is a nonlinear differential operator, $\z$ is a set of parameters for which we solve the inverse problem, $\w$ is an extra set of model parameters for which we would like to learn the forward and inverse maps\footnote{The inverse problem is defined here w.r.t. $\mathbf{z}$, not $\mathbf{w}$.}. Similarly, $\cB_\w$ is the boundary operator. For the forward problem, consider the problem of learning a forward parametric emulator $f_\alpha: \z, \w, x \rightarrow u(x)$ where $\alpha$ denotes the parameters of the emulator. This can be informally formulated as the approximation problem of finding $\alpha_\star$ s.t.
\begin{align}
    \cG^\w_\z \circ f_{\alpha_\star}(\z,\w)(x) \approx 0,
\end{align}
given the boundary conditions, following \eqref{eq:PDE}. This can be formulated into an optimization problem or can inform a probabilistic model \cite{kaltenbach2023semi, gao2022physics, zhao2022learning}. In the context of PDEs, the residual is defined as $r = \cG^\w_\z(u)(x)$. When $r(x) =0$ then $u(x)$ is the solution to the PDE \eqref{eq:PDE}.

Another important aspect to consider is the boundary conditions. In general, there are many ways to enforce boundary conditions \cite{raissi2019physics}. In this work, we use the hard enforcement method. 
We can enforce hard boundary conditions for Dirichlet problems \cite{rao2021physics, sukumar2022exact} with a linear transformation of the solution field
\begin{align}
    u(x) = B(x) + D(x)N(x),
\end{align}
where $B(x)$ is an arbitrary function that satisfies the boundary conditions and $D(x) = 0 $ for $x\in\partial\Omega$, and $D(x)\neq0$ for $x\in\Omega$. Other formulations exist for mixed enforcement of boundary conditions as well as for more complicated boundary conditions.

\subsection{Physics Informed ML for Inverse Problems}
We now consider the problem of fitting a deterministic parametric inverse emulator denoted as
$h_\beta: u(x),\w \rightarrow \z$ where we seek $\beta_\star$ s.t.
\begin{align}
     h_{\beta_\star}(f_{\alpha_\star}(\z,\w,\bX), \w) \approx \z ,
\end{align}
for any subset $\bX$ of the domain $\Omega$. Similarly to the forward case, this can be done using an optimization or a probabilistic formulation \cite{vadeboncoeur2022deep}. We note that the inverse emulator approximation problem proposed here relies on a trained forward emulator and is free from classical numerical solvers. In what follows, we develop a comprehensive probabilistic framework to tackle such problems in a principled way.

\section{Model Derivation}\label{sec:model_derivation}
In this section, we derive the training objective for two cases of our model: physics informed model in Sec.~\ref{sec:phys_only} and physics and data informed model in Sec.~\ref{sec:phys_data}. In Appendix~\ref{app:derivation_data_intractable} we derive a model for incorporating observations of nonlinear transformations of solution fields and noisy input parameters  e.g. measurements of drag coefficients. 
\subsection{Physics Informed Probabilistic Framework}\label{sec:phys_only}
\textbf{Function Space Model:} We begin the derivation of the physics informed probabilistic model by introducing all distributions of interest with a random field over our solution field. Our hierarchical probabilistic model is defined as
\begin{align}
    r|u, \z, \w &\sim \cGP(\cG^\w_\z(u)(x), k_r(x, x';u, \z, \w)), \label{eq:residual} \\
        \z | u, \w &\sim \NPDF(\bmu_{\beta}(u), \bSigma_\beta(u, \w)),\label{eq:beta_net} \\
            u &\sim \mathcal{GP}(\mu_u(x), k_u(x, x')), \label{eq:prior_functional_u}\\
    \w &\sim \NPDF(\bmu_\w, \bSigma_\w).\label{eq:prior_w}
\end{align}
In this model, Eq.~\eqref{eq:residual} defines a probability distribution over the residual and informs  the model with physics. Eq.~\eqref{eq:beta_net} is the \textit{inverse emulator} to be learned during training for the inverse map. Finally, \eqref{eq:prior_functional_u} and \eqref{eq:prior_w} define the priors over $u$ and $\w$. Leveraging the GP view allows us to choose from several possible choices of kernels for our different distributions. Some of these include
\begin{align}
    k_\theta(x,x')  &= \epsilon_\theta\delta_{x,x'},\label{eq:kernelFixWhiteNoise}\\
    k_\theta(x,x')  &= \lambda_\theta(x)\delta_{x,x'},\label{eq:kernelHeterosceWhiteNoise}\\
    k_\theta(x, x') &= \lambda_\theta(x)\delta_{x,x'} + \langle V_\theta(x), V_\theta(x') \rangle. \label{eq:kernelLR}
\end{align}
Here $\theta$ denotes learnable functions or parameters. The kernels in \ref{eq:kernelFixWhiteNoise}--\eqref{eq:kernelLR} correspond to a fixed white noise process, a heteroscedastic white noise process (diagonal kernel), and a degenerate deep low-rank covariance matrix (also known as the left Gram matrix \cite{rahman2022using, williams2006gaussian}), respectively. We elaborate on \eqref{eq:kernelLR} in Appendix~\ref{app:low_rank_covriance}. These kernels are chosen for their favorable scalability properties. \\

\textbf{Function Space Variational Family:} We next introduce the variational family in function space (which will be discretized later). For this, we define
\begin{align}
    u | \z, \w  &\sim  \cGP(\mu_{\alpha}(x;\z, \w), k_{\alpha}(x, x';\z, \w)), \label{eq:func_space_variational_u} \\
    \z &\sim \NPDF(\bmu_\z, \bSigma_\z) \label{eq:prior_z}
\end{align}
where \eqref{eq:func_space_variational_u} is a flexible variational family parameterized by a neural network to learn the \textit{forward emulator}. Eq.~\eqref{eq:prior_z} define the prior in the variational family while the variational distribution over $\w$ is the same as the prior in Eq.~\eqref{eq:prior_w}. \\

\textbf{Discretization through Conditioning:} To obtain a tractable algorithm, we condition all distributions on a set of $N$ (grid) points $\bX \subset \Omega$ in the domain of the PDE. We achieve this by assigning a joint probability measure on $\Omega^{\otimes N}$ which we denote as $p(\bX)$. Sampling from this measure and conditioning on the sample effectively discretizes the Gaussian processes. Given the sample, we can discretize our distributions through conditioning on $\bX$ as in \citet{cressie2021spatial} and convert distributions defined on the function space into multivariate normal distributions (MVN) \cite{rudner2021tractable}. More precisely, through conditioning on $\bX$, we can convert the infinite dimensional model given in \eqref{eq:residual}--\eqref{eq:prior_w} into a conditional finite-dimensional model where
\begin{align}
    p(\r|\ubf, \z, \w , \bX) &= \NPDF(\cG^\w_\z(\ubf), K_r(\bX, \bX;\ubf, \z, \w)), \label{eq:conditioned_residual} \\
    p_\beta(\z|\ubf, \w, \bX) &= \NPDF(\bmu_{\beta}(\ubf, \bX, \w), \bSigma_\beta(\ubf, \bX, \w)), \label{eq:conditioned_beta_dist}\\
    p(\ubf | \bX) &= \NPDF(\bmu_u(\bX), K_u(\bX, \bX)) \label{eq:conditioned_u_dist}
\end{align}
where $\ubf = u(\bX)$, and $\r \in \Rb^{N}$ is the discretized residual (a vector). For the conditional residual \eqref{eq:conditioned_residual} we choose a white noise kernel \eqref{eq:kernelFixWhiteNoise} and the mean function is given by the PDE evaluated at locations $\bX$. The conditional distribution for  $\z$ \eqref{eq:conditioned_beta_dist} has mean and covariance given as the output of a neural network with learnable $\beta$ parameters. We call this network the ``$\beta$-Net'' and it relates the probability density of the $\z$ parameter to the solution field $\ubf$ and the parameters $\w$ given a partition of the domain. The distribution \eqref{eq:conditioned_u_dist} is chosen in practice to be uninformative so that the model predictions are purely influenced by the PDE residual.
The prior on $p(\w)$ is the same as \eqref{eq:prior_w} since it is independent of the grid. Finally, our joint model can be factorized as
\begin{align}
p(\r,&\ubf, \z, \w,\bX) = \\
    &p(\r|\ubf, \z, \w , \bX)p_\beta(\z|\ubf, \w,\bX)p(\ubf|\bX)p(\w) p(\bX) \nonumber.
\end{align}
 From this joint model we are interested in the marginal residual which can be obtained by integrating out all other variables (including the grid) as
\begin{align}
    p(\r) &= \int p(\r, \ubf, \z, \w, \bX)\,\md\ubf\,\md\z\,\md\w\,\md\bX.
\end{align}
We know that the PDE is solved if $\r = \b0$, hence our aim will be to maximize the marginal probability $p(\r = \b0)$. For this, we discretize our variational approximation in \eqref{eq:func_space_variational_u}--\eqref{eq:prior_z} by conditioning on the grid $\bX$. In addition, we can convert the infinite dimensional variational approximation in \eqref{eq:func_space_variational_u} to a finite dimensional one by conditioning on $\bX$ as
\begin{align}
    \hspace{-0.05cm}\label{eq:conditioned_alpha_dist}
    q_\alpha(\ubf | \z, \w, \bX)  = \NPDF(\mu_{\alpha}(\bX;\z, \w), K_{\alpha}(\bX, \bX;\z, \w))
\end{align}
where \eqref{eq:conditioned_alpha_dist} is a conditional neural process with $\alpha$ learnable weights that relates the probability density of the solution field $\ubf$ to the two sets of parameters $\z$ and $\w$. This neural process is given by the $\alpha$-Net which is the forward emulator. In the experiments we alternate between a heteroskedastic white noise kernel \eqref{eq:kernelHeterosceWhiteNoise}, and a low-rank kernel \eqref{eq:kernelLR}.
Finally, our joint variational approximation can be written as
\begin{align}
    q(\ubf, \z, \w, \bX) = q_\alpha(\ubf|\z, \w, \bX) q(\z) p(\w) p(\bX).
\end{align}

We then set the virtual observable  $\r = \b0$ \cite{rixner2021probabilistic, vadeboncoeur2022deep}. Using Jensen's inequality we write out the evidence lower bound
\begin{align}
    &\log p(\r = \b0)\geq\cF(\alpha, \beta) \\
    & = \int \log \frac{p(\r=\b0|\ubf, \z, \w, \bX)p_\beta(\z|\ubf, \w, \bX)p(\ubf|\bX)}{q_\alpha(\ubf|\z, \w, \bX)q(\z)}\nonumber\\
    &\times q_\alpha(\ubf|\z, \w, \bX)q(\z)p(\w) p(\bX)\,\md\ubf\,\md\z\,\md\w\,\md\bX\nonumber.
\end{align}
The most important distinction between this objective and all other methods known to us is the marginalization of the spatio-temporal domain through conditioning on random partitions of the domain. This objective can be computed as an expectation of the form
\begin{align}\label{eq:elbo_phys}
    &\cF(\alpha, \beta)= \\
    &\Eb_{\ubf, {\z}, {\w}, {\bX}} \left[ \log \frac{p(\r=\b0|\ubf, \z, \w, \bX)p_\beta(\z|\ubf, \w, \bX)p(\ubf|\bX)}{q_\alpha(\ubf|\z, \w, \bX)q(\z)}\right].\nonumber
\end{align}
This expectation can be approximated with Monte Carlo integration. 
Every Monte Carlo sample evaluation requires a newly sampled set of collocation points akin to methods of variational inference in function space \cite{burt2020understanding, sun2018functional}. In Alg.~\ref{alg:rgnp_alg} we write out the procedure for training a physics informed RGNP. A similar algorithm is used for the data and physics case where we then add a mini-batched data likelihood term in $\cF^N(\alpha, \beta)$. The RGNP update starts in the latent space and maps a sampled parameter into the solution field, from this
proposed solution field we compute a physics residual and possibly a data likelihood, and then map this proposed solution field back to the parameters space. Because of this fundamental difference in the construction of the ELBO, our method can work in the complete absence of solution field data. Furthermore, our method generates solution fields of PDEs in function space, i.e. they can be evaluated anywhere in the domain.

\subsection{Physics and Data Informed Model Derivation}\label{sec:phys_data}
When developing physics emulators to be used in practice, we may have noisy observation of real world physics behaviours. Data of this kind can be of great use when developing better calibrated and more accurate models \cite{zhong2022pi, takeishi2021physics}. In the current state of the art there is a lack of methods that map parameters to solution fields adjusted with data in a statistically principled manner. Many methods which do incorporate data in a Bayesian manner then map parameters to the observation space, which may not be the desired output. In this section we derive a model for direct observation of a noisy solution field with deterministic inputs. The Bayesian approach is to pose a model of the form
\begin{align}
    \y_D = G(\z,\w, \bX) + \sigma_n\ebf, \quad \ebf\sim \NPDF(\b0, \bI),
\end{align}
where $\y_D$ is the data and $\sigma_n$ is the observational noise, and $G$ is the mapping from parameters to solution field $\ubf$. For our method we pose this mapping $G(\z,\w,\bX)$ to be our stochastic forward model given the $\alpha$-Net,
\begin{align}
    \y_D^i &= \bmu_\alpha(\z_D^i, \w_D^i, \bX_D^i) \\
    &+ \bK_{\alpha}(\bX_D^i, \bX_D^i;\z_D^i, \w_D^i)^{\frac{1}{2}}\ebf_2 + \sigma_n \ebf_1\nonumber
\end{align}
where $\ebf_1, \ebf_2  \sim \NPDF(\b0, \bI)$ and $i$ indexes the observation set. In our framework we then jointly learn the forward model through the $\alpha$-Net while adjusting the predictions to match the observed noisy dataset taking into account the relevant uncertainties. Our full joint model can then be written as
\begin{align}
    &\log p(\r=\b0, \y_D|\z_D,\w_D,\bX_D)\geq \cF(\alpha, \beta)\\
     &= \sum_{i}^N \log p(\y^i_D|\z^i_D, \w^i_D, \bX^i_D) \nonumber\\
     &+\Eb_{\ubf, {\z}, {\w}, {\bX}} \left[ \log \frac{p(\r|\ubf, \z, \w, \bX)p_\beta(\z|\ubf, \w, \bX)p(\ubf|\bX)}{q_\alpha(\ubf|\z, \w, \bX)q(\z)}\right].\nonumber
\end{align}
The lower bound on the marginal likelihood of the zero residual and the observed data requires the evaluation of the entire dataset at every iteration. For computational efficiency, we can replace the summation over the entire dataset with a mini-batch approximation as
\begin{align}\label{eq:elbo_phys_data}
    &\cF(\alpha, \beta) \approx \frac{N}{|M|}\sum_{i\in M} \log p(\y^i_D|\z^i_D, \w^i_D, \bX^i_D)\\
    &+ \Eb_{\ubf, {\z}, {\w}, {\bX}} \left[ \log \frac{p(\r|\ubf, \z, \w, \bX)p_\beta(\z|\ubf, \w, \bX)p(\ubf|\bX)}{q_\alpha(\ubf|\z, \w, \bX) q(\z)}\right].\nonumber
\end{align}
The first term of the ELBO can be efficiently evaluated and the second term can be approximated using a finite sample size through a Monte Carlo approximation of the expectation.



\subsection{Grid Invariant Inversion Networks}\label{sec:grid_inv_net}
Central to this new framework is the use of random collocation grids which are sampled every iteration. This poses unique challenges for the inversion network. Convolutional neural networks (CNNs) require information to be given on a fixed uniform grid and are thus unsuitable for our task. A natural solution for passing information originating on random grids to a CNN would be to use a kernel interpolation method such as the Nadaraya–Watson kernel estimator  \cite{cai2001weighted} to interpolate points at random locations to fixed input locations. 
However, to capture information from different length scales, a very fine projection grid is required. 
Such grids are computationally expensive and scale as $\cO((nm)^{d})$, where $m$ is the number of points on the projection grid. Borrowing ideas from \citet{li2020fourier} and kernel feature space methods \cite{scholkopf1999input}, we develop a scalable alternative to fine interpolation grids. We first project the spatially dependent inputs $u(x)\in\Rb^{d_s}$ to a learned higher dimensional space through a small fully connected neural network $P(u(x), x)= v(x)$ where $ v(x)\in\Rb^{d_v}$ and $x\in\Omega\subset \Rb^d$. We then project each dimension of $v(x)$ at random locations $x$ onto its own fixed location coarse grid $x^*$ through the Nadaraya-Watson estimator
\begin{equation}
    I_k(v(x), x^*) = \frac{\sum_i \phi_k(x^*,x_i)v(x_i)}{\sum_i\phi_k(x^*,x_i)},
\end{equation}
where each of the $d_v$ kernels has its own learnable length scale initialized at several different orders of magnitude. A visual representation of this can be seen in the Appendix, Fig.~\ref{fig:signal_feature_space_representation}. Each coarse grid can then represent information at different lengthscales and the complexity grows as $\cO(d_v(nm')^{d})$ where $m'$ is the new grid density of the interpolation grid which can now be chosen to be arbitrarily coarse and in practice is chosen as coarse as a 10 points/dimension lattice. We can choose the kernel $\phi_k(\cdot, \cdot)$ to be any number of distance measuring functions with learnable parameters such as RBF, Matern, etc. We then define the grid invariant convolutional network (GICNet) as
\begin{align}
    G(u,& \w, x) = \text{Conv}(\cdot, \w) \circ I_{1:d_v}(\cdot, x^*) \circ P(\cdot, x) \circ u(x)\nonumber
\end{align}
where the projection layer $P(u(x), x)$, the interpolation layer $I(v(x), x)$ along with the kernel $\phi(x^*,x)$ lengthscale and the convolutional and fully connected layers are jointly trained.
Other architectures than a convolutional network can be used for mapping the $d_v$ intermediary function representations to the output PDE parameters. Architectures of interest include Fourier Neural Operator layers~\cite{li2020fourier} and Wavelet Neural Operator layers~\cite{tripura2023wavelet} as well as fully connected layers. Each of these may bring certain information processing advantages for PDE inversion networks to be explored in future work.

\begin{algorithm}[tb]
  \caption{Pseudocode for RGNP\label{alg:rgnp_alg}}
  \begin{algorithmic}
    \STATE Initialise: $\alpha_0$, $\beta_0$, $T$ (number of iterations), $N$ (number of Monte Carlo samples), and choose $p(\z)$, $p(\w)$, $p(\bX)$.
    \FOR{$t=1, \ldots,  T$}
    \FOR {$i = 1, \ldots, N$}
      \STATE Sample $\bX^{(i)} \sim p(\bX)$
      \STATE Sample $\z^{(i)} \sim p(\z)$
      \STATE Sample $\w^{(i)} \sim p(\w)$
      \STATE Sample $\ubf^{(i)} \sim q_{\alpha_{t-1}}(\ubf | \z^{(i)}, \w^{(i)}, \bX^{(i)})$
     \ENDFOR
      \STATE Compute $\cF^N(\alpha, \beta)$ using Monte-Carlo.
      \STATE $(\alpha_t, \beta_t) \gets \textnormal{ADAM}(\alpha_{t-1}, \beta_{t-1}, \cF^N(\alpha, \beta))$
    \ENDFOR
  \end{algorithmic}
\end{algorithm}

\section{Experiments}\label{sec:experiments}
In this section, we test our method on a series of PDEs and compare our method to alternative approaches. We outline the setup for the three parameterized PDEs used for comparisons, namely a 1D nonlinear Poisson problem, the Burgers equation, and a non-stationary lid-driven cavity flow Navier-Stokes problem. The chosen testing metrics are the mean normalized squared error (see Appendix, \eqref{eq:mnse}) averaged over 1000 independent samples of $\z, \w$ drawn from their priors (100 samples in the Navier-Stokes examples) solved using FEniCS \cite{logg2012automated}. We set the $\epsilon_r$ value in the residual kernel to a value of $10^{-2}$ other than for the Burgers example as explained in Sec.~\ref{sec:burgers_equation}.
\subsection{The PDEs}
In this section, we describe the PDEs used in the testing of the method along with the relevant boundary conditions and their parametrizations.
\begin{table*}[t]
\centering
\caption{Comparisons of Physics Informed Models}
\vskip 0.15in
\begin{center}
\begin{small}
\begin{sc}
\begin{tabular}{lcccccr}
\toprule
     Method          & n. coll & MNSE $u$     & MNSE $\z$    & $ u \text{ in } 2\sigma$ & $ \z \text{ in } 2\sigma$ & time \\
\midrule
    \textbf{NL Poisson 1D}\\
\midrule
\textbf{rgnp-d}         &  30  & $\mathbf{9.21\cdot10^{-5}}$  $\mathbf{\pm8.66\cdot10^{-4}}$ & $1.48\cdot10^{-2}$  $\pm2.41\cdot10^{-2}$ & $\mathbf{95.6\%}$ & $99.7\%$ & 6.31 \\
\textbf{rgnp-lr} $\alpha~2, \beta~1$        &  30  & $2.63\cdot10^{-4}$  $\pm1.88\cdot10^{-3}$ & $6.72\cdot10^{-2}$  $\pm5.96\cdot10^{-2}$ & $94.7\%$ & $68.1\%$ & 6.19 \\
pddlvm                 &  30  & $1.69\cdot10^{-4}$ $\pm1.36\cdot10^{-3}$ & $\mathbf{8.10\cdot10^{-3}}$  $\mathbf{\pm1.36\cdot10^{-3}}$ & $92.3\%$ & $\mathbf{95.2\%}$ & 4.83 \\
FFNet \& GICNet        &  30  & $1.90\cdot10^{-4}$ $\pm2.04\cdot10^{-3}$ & $6.75\cdot10^{-1}$  $\pm1.57\cdot10^{-0}$ &   --       &  -- & 5.06     \\
DeepONets \& k-i.   &  30  & $6.62\cdot10^{-4}$ $\pm3.41\cdot10^{-3}$ & $5.58\cdot10^{-2}$  $\pm4.62\cdot10^{-2}$ &   --       &  -- & 3.31     \\
DeepONets \& k-i.   &  100 & $6.62\cdot10^{-4}$ $\pm3.95\cdot10^{-3}$ & $3.66\cdot10^{-2}$  $\pm2.48\cdot10^{-2}$ &   --       &  -- & 5.36    \\
DeepONets \& k-i.   &  300 & $7.69\cdot10^{-4}$ $\pm3.90\cdot10^{-3}$ & $2.58\cdot10^{-2}$  $\pm2.58\cdot10^{-2}$ &   --       &  -- & 11.34    \\
\midrule
    \textbf{Burgers}\\
\midrule
\textbf{rgnp-d}                      &  225   & $1.63\cdot10^{-4}$  $\pm1.58\cdot10^{-4}$ & $9.05\cdot10^{-3}$  $\pm1.73\cdot10^{-2}$ & $99.9\%$ & $\mathbf{100.0}\%$ & 73.56 \\
\textbf{rgnp-lr} $\alpha~2, \beta~0$ &  225   & $\mathbf{9.51\cdot10^{-5}}$  $\mathbf{\pm9.73\cdot10^{-5}}$ & $\mathbf{8.04\cdot10^{-3}}$  $\mathbf{\pm1.66\cdot10^{-2}}$ & $99.9\%$ & $\mathbf{100.0\%}$ & 73.45\\
pddlvm                              &  225   & $2.70\cdot10^{-1}$  $\pm2.12\cdot10^{-1}$ & $5.32\cdot10^{-2}$  $\pm6.49\cdot10^{-2}$ & $\mathbf{97.7\%}$ & $63.0\%$ & 59.02 \\
FFNet \& GICNet                     &  225   & $7.11\cdot10^{-1}$  $\pm1.70\cdot10^{-1}$ & $1.19\cdot10^{-1}$  $\pm1.75\cdot10^{-1}$ & --     & --       &  66.45 \\
DeepONets \& k-i.                &  225   & $9.44\cdot10^{-1}$  $\pm4.68\cdot10^{-1}$ & $8.66\cdot10^{-2}$  $\pm1.45\cdot10^{-1}$ & --     & --       & 36.00 \\
DeepONets \& k-i.                &  900   & $4.16\cdot10^{-1}$  $\pm2.07\cdot10^{-1}$ & $9.79\cdot10^{-1}$  $\pm7.15\cdot10^{-1}$ & --     & --       & 105.61 \\
DeepONets \& k-i.                &  2.5k  & $9.65\cdot10^{-2}$  $\pm1.05\cdot10^{-1}$ & $5.04\cdot10^{-1}$  $\pm2.32\cdot10^{-1}$ & --     & --       & 281.14 \\
DeepONets \& k-i.\footnotemark
                                   &  10k & $3.71\cdot10^{-3}$  $\pm8.22 \cdot10^{-3}$ & $2.01\cdot10^{-2}$  $\pm5.35\cdot10^{-2}$ & --     & -- & 284.90   \\
\bottomrule
\end{tabular}
\end{sc}
\end{small}
\end{center}
\vskip -0.1in
\label{tbl:comparisonsPhysOnly}
\end{table*}
\subsubsection{Nonlinear Poisson 1D}
The first testing setup is a nonlinear 1D Poisson problem. The variable $\z$ which we want to learn in the inverse problem are coefficients to a Chebyshev expansion describing the diffusion field. The variable $\w$ over which we are marginalizing is a scalar representing a constant forcing over the domain. We write out the PDE $\cG_\z^\w(u)(x)$ is given as
\begin{align}
    &\frac{\partial}{\partial x}\left(k(u,x)\frac{\partial u(x)}{\partial x}\right) - w = 0,  \\
    &k(u,x) = \log\left( 1 + \exp\left(u(x)\sum_{i=0}^{n_{z}} z_i\phi_i(x)\right) \right) + 0.1\nonumber.
\end{align}
for $\Omega = [-1, 1]$, where the Dirichlet boundary conditions are $x(-1)= 0$, $x(1) = 0$. Furthermore, we enforce boundary conditions in an exact manner with,
\begin{align}\label{eq:exact_boundary_sampling}
    \bar{\ubf} &= B(\bX) + D(\bX) \ubf, \quad \ubf \sim q_\alpha(\ubf|\z,\w,\bX)
\end{align}
where $B(x)$ captures the boundary conditions. 
The prior distributions are $p(z_i) = \cU(-1,1)$ and $p(w) = \cU(1,2)$.
\begin{table*}[t]
\label{tbl:comparisonsPoisson1D}
\centering
\caption{Comparisons of Physics and Noisy Data Informed Models}
\vskip 0.15in
\begin{center}
\begin{small}
\begin{sc}
\begin{tabular}{lccccccr}
\toprule
     Method          & n. coll & MNSE $u$    & MNSE $\z$     & $ u \text{ in } 2\sigma$ & $ \z \text{ in } 2\sigma$ & time \\
\midrule
    \textbf{NL Poisson 1D}\\
\midrule
     \textbf{rgnp-d}         &  30  & $\mathbf{1.59\cdot10^{-5}}$  $\pm5.26\cdot10^{-5}$ & $\mathbf{5.45\cdot10^{-3}}$  $\mathbf{\pm6.25\cdot10^{-3}}$ & $97.2\%$ & $99.7\%$ & 6.21 \\
     \textbf{rgnp-lr} $\alpha~2, \beta~1$        &  30  & $3.56\cdot10^{-5}$  $\pm2.48\cdot10^{-4}$ & $3.43\cdot10^{-2}$  $\pm7.40\cdot10^{-2}$ & $98.6\%$ & $62.9\%$ & 6.47 \\
     pddlvm                 &  30  & $1.67\cdot10^{-5}$  $\mathbf{\pm3.60\cdot10^{-5}}$ & $7.26\cdot10^{-3}$  $\pm7.23\cdot10^{-3}$ & $\mathbf{95.3\%}$ & $\mathbf{97.6\%}$ & 4.23 \\
     FFNet \& GICNet        &  30  & $2.85\cdot10^{-4}$  $\pm8.76\cdot10^{-4}$ & $4.36\cdot10^{-1}$  $\pm5.76\cdot10^{-1}$ & -- & -- & 5.59 \\
     DeepONets \& k-i. &  30  & $1.55\cdot10^{-4}$  $\pm2.77\cdot10^{-4}$ & $9.70\cdot10^{-2}$  $\pm8.99\cdot10^{-2}$ & -- & -- & 3.94 \\
     DeepONets \& k-i. &  100 & $2.17\cdot10^{-4}$  $\pm2.86\cdot10^{-4}$ & $1.29\cdot10^{-1}$  $\pm1.52\cdot10^{-1}$ & -- & -- & 5.67 \\
     DeepONets \& k-i. &  300 & $3.27\cdot10^{-4}$  $\pm4.44\cdot10^{-4}$ & $6.97\cdot10^{-2}$  $\pm4.56\cdot10^{-1}$ & -- & -- & 11.70 \\
\bottomrule
\end{tabular}
\end{sc}
\end{small}
\end{center}
\vskip -0.1in
\end{table*}
\subsubsection{Burgers Equation}\label{sec:burgers_equation}
The second PDE used to test the method is the parametric Burgers equation with 2 spatio-temporal dimensions (one space, one time). Here the $\z$ parameters control the scaling of the nonlinear and the diffusive term and the $\w$ scalar modifies the parametric initial conditions. This allows us to learn forward and inverse mappings for a continuum of initial conditions of the given form. The PDE $\cG_\z^\w(u)(x)$ is defined as
\begin{align}
    &\frac{\partial u(x,t)}{\partial t} + z_1 u(x,t) \frac{\partial u(x,t)}{\partial x} - z_0  \frac{\partial^2 u(x,t)}{\partial x^2} = 0,
\end{align}
and the boundary and initial conditions are
\begin{align}
    &u(-1, t) = u(1,t) = 0, \\
    &u(x, 0) = \sin(2\pi wx)\sin(\pi x),
\end{align}
for a set domain given as $\Omega [-1,1], \quad 0 \leq t \leq 1$ \cite{duffin2022statistical}. The priors are: $p(z_0) = \cU(10^{-2},10^{-1})$, $p(z_1)=\cU(0.5, 1)$, $p(w)=\cU(0.5, 2)$. As in the nonlinear Poisson example we enforce boundary conditions in an exact manner with \eqref{eq:exact_boundary_sampling}. In this example, we make the standard deviation of the residual kernel value ($\epsilon_\theta$ in \eqref{eq:kernelFixWhiteNoise}) a learnable parameter which drastically increases stability while maintaining accuracy. It converges to a value in the range of $10^{-1}$. 

\subsubsection{Navier-Stokes Equations}
The final PDE used to test the method is the incompressible Navier-Stokes non-stationary lid-driven cavity flow. This is a setup for studying fundamental aspects of confined fluid-flows \cite{botella1998benchmark}. The solution field is a 3-dim. vector field defined as the velocity and pressure $[u, p]^\top$, where $u = [u_1, u_2]^\top$ are the horizontal and vertical velocities.
We do not define a $\w$ variable for these experiments. The equations for $\cG_\z^\w(u)(x)$ describing the dynamics are given by
\begin{align}
    &z_1\frac{\partial u}{\partial t} + z_1 u\cdot\nabla u + \nabla p - z_2\nabla^2 u = 0,\\
    &\nabla \cdot u = 0
\end{align}
over a square domain, $\Omega(x,y,t)$ between 0 and 1.
We define a non-stationary boundary condition on the top edge of the square domain as
\begin{align}
    &u_1(x,1,t) = (1-(2x-1)^6) t,
\end{align}
and all other boundary conditions of $u_1, u_2$ are zero.
The $\alpha$ and $\beta$ distributions are then defined as 
\begin{align}
    u_1, u_2, p | \z  &\sim  \cGP(\mu_{\alpha}(x;\z), k_{\alpha}(x, x';\z)), \\
    \z | u_1, u_2, p &\sim \NPDF(\bmu_{\beta}(u_1,u_2,p), \bSigma_\beta(u_1, u_2,p)).
\end{align}
The $z_1$ variable corresponds to the fluid density and the $z_2$ parameter corresponds to the dynamic viscosity. Here $p(z_1)=\cU(0.8, 1)$ and $p(z_2)=\cU(0.1, 1)$. For this example, we enforce exact boundary conditions with \eqref{eq:exact_boundary_sampling} but enforce a soft divergence as an extra residual appended in the residual vector  (scaled $10\times$).
\begin{figure}
    \vskip 0.2in
    \begin{center}
    \centerline{\includegraphics[width=0.95\columnwidth]{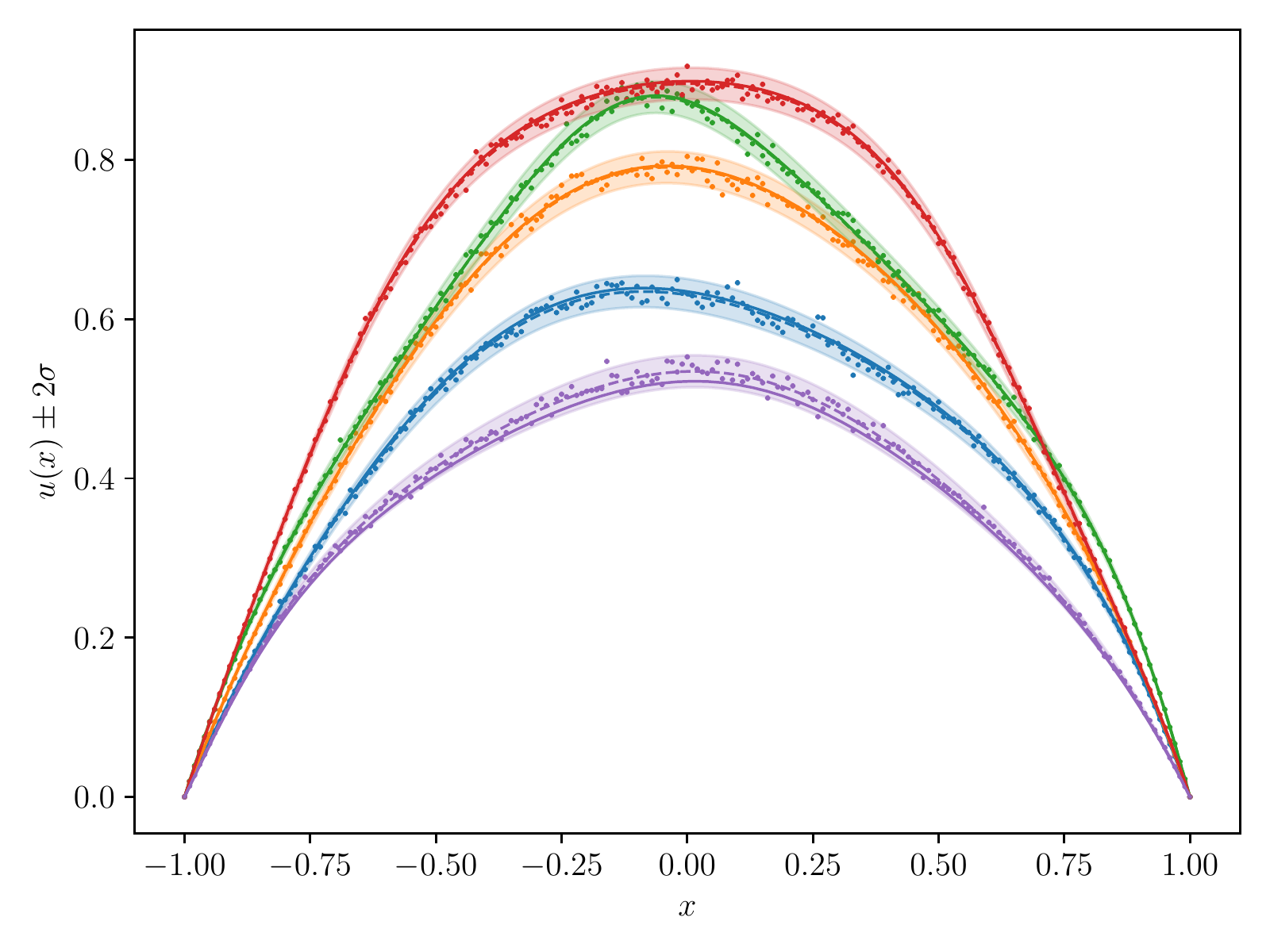}}
    \caption{Sample results from the Nonlinear Poisson 1D setup for the diagonal model given by the $q_\alpha(\ubf|\z,\w,\bX)$ distribution. Here, the solid line is the FE solution, the dashed line is the mean estimate, and the scatter points are the random samples; the shaded area is the 2$\sigma$ uncertainty. }
    \label{fig:rgnp-d-poisson1D}
    \end{center}
    \vskip -0.2in
\end{figure}
\footnotetext{We used a reduced batched size of 5 collocation sets because of memory constraints. The same number of gradient updates were used hence the similar run-times for the 2.5k and 10k examples.}
\subsection{Comparisons of Physics Informed Model}
\begin{figure}[t]
    \centering
    \vskip 0.2in
    \subfigure[FE solution]{
        \includegraphics[width=0.48\columnwidth]{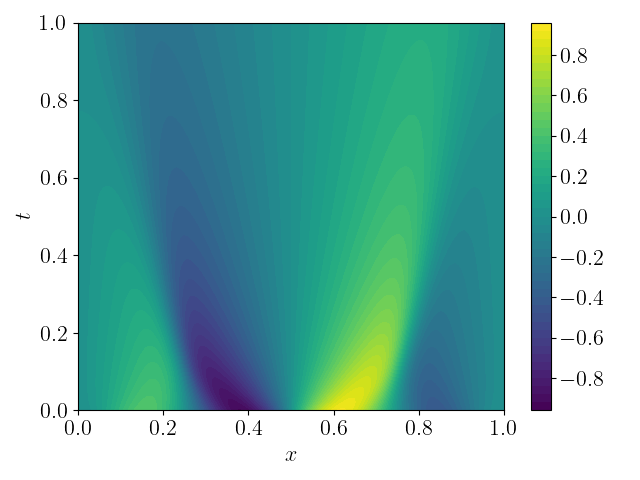}}    
    \subfigure[mean-$\alpha$]{
        \includegraphics[width=0.48\columnwidth]{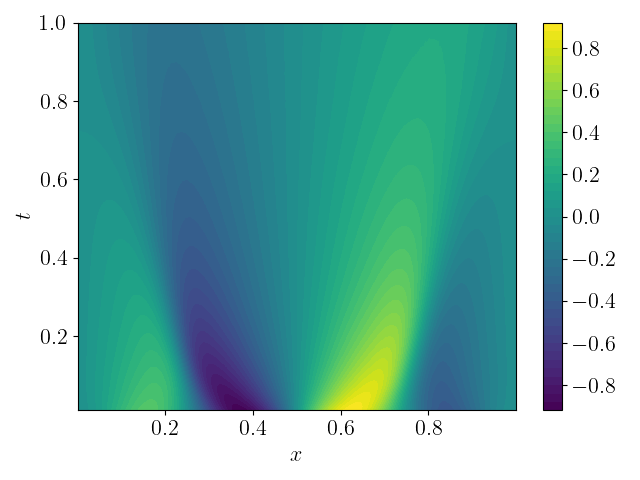}}   
    \\
    \subfigure[sample-$\alpha$]{
        \includegraphics[width=0.48\columnwidth]{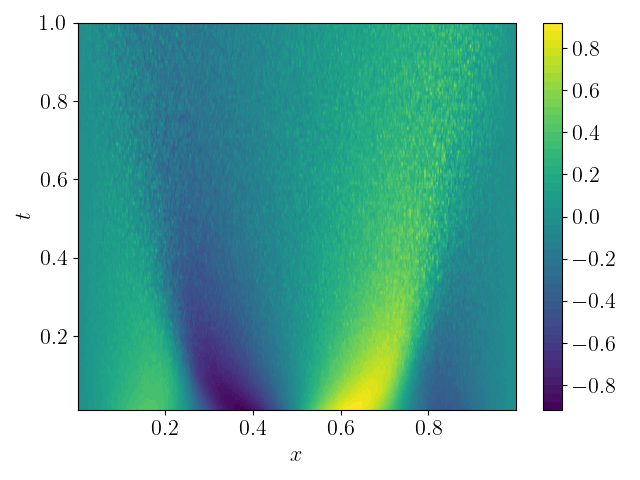}}    
    \subfigure[stddev-$\alpha$]{
        \includegraphics[width=0.48\columnwidth]{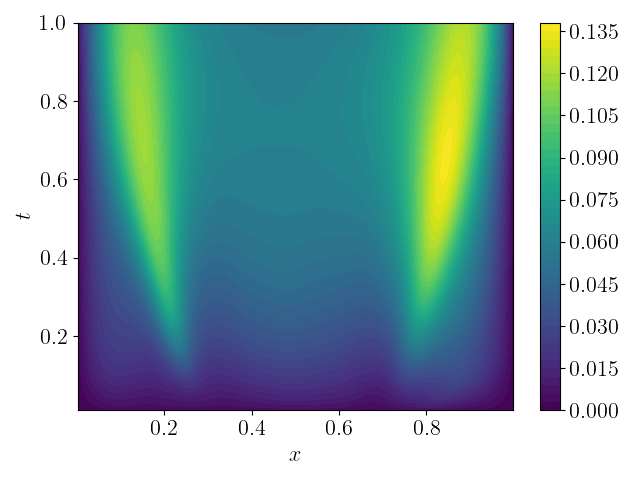}}
    \caption{Sample result from the $\alpha$-Net with a diagonal kernel for the Burgers equation. We show the FE solution, the mean estimate, the standard deviation field, and a sample from $q_\alpha(\ubf|\z,\w,\bX)$.}
    \label{fig:rgnp-d-burger}
    \vskip -0.2in
\end{figure}

In this section we describe the physics informed comparisons for the nonlinear Poisson and Burgers equation. We compare our diagonal and low-rank covariance methods (RGNP-D, RGNP-LR) to a fixed grid PDDLVM \cite{vadeboncoeur2022deep}, a Fourier Feature Net (FFNet) forward emulator \cite{wang2021eigenvector} with a GICNet inversion network on a fixed grid, and a physics informed DeepONet \cite{wang2021learning} with a test-time kernel interpolation (K-I) layer with a convolutional neural net on a fixed grid. Methods other than ours rely on creating a dataset of 1k input pairs of $\z, \w$ variables for the Poisson problem, and 10k samples for the Burgers example. Table \ref{tbl:comparisonsPhysOnly} summarizes the results for all methods for the two PDE setups. We report the MNSE as in \eqref{eq:mnse} and its standard deviation over the 1000 testing samples. We also report the percentage of predictions within $2\sigma$ of the ground truth solution and the total training times in minutes. Test time inference is in the order of $10^{-2}-10^{-3}$ seconds. The column labeled ``N. COLL'' denotes the number of collocation points used at every
residual computation step. We use a batch size of 50 samples for the Poisson problem, and a batch size of 25 for the Burgers equation (this is reduced to a batch size of 5 for the 10k collocation DeepONet \& kernel-interpolation case because of excessive memory consumption). We train the Poisson problem for 20k gradient updates and we train the Burgers setup for 80k gradient update steps. All learning is done using the Adam optimizer \cite{kingma20153rd} with a decaying learning rate.
In Fig.~\ref{fig:rgnp-d-poisson1D} we show an example output from the $\alpha$-Net for the solution field of the nonlinear Poisson 1D problem. In Fig.~\ref{fig:rgnp-d-burger} we show example outputs for a solution field for the diagonal model from the $\alpha$-Net. The main reason for the gain in performance of the probabilistic model is due to the stochastic treatment of the collocation points. By treating the collection of collocation points as random variables to be marginalized through Monte Carlo integration we obtain domain averaged residuals which proves to be advantageous when learning parametric physics.
\begin{table*}[t]
\label{tbl:resultsNSlid}
\centering
\caption{Results for Physics informed models applied to Navier-Stokes lid-driven cavity flow}
\vskip 0.15in
\begin{center}
\begin{small}
\begin{sc}
\begin{tabular}{lccccccccr}
\toprule
     Method          & n. coll & MNSE $u_1$   & MNSE $\z$  & $ u_1 \text{ in } 2\sigma$ & $ \z \text{ in } 2\sigma$ & time \\
                     &         & MNSE $u_2$   &            & $ u_2 \text{ in } 2\sigma$ &                           &      \\
\midrule
     \textbf{rgnp-d}                      & 512 & $2.37\times 10^{-2}$ $\pm 1.16\times 10^{-2}$ &  $6.12\times 10^{-3}$ $\pm 6.00\times 10^{-3}$ &  $38.3\%$ & $66.5\%$ & $297.6$\\
                                         &     & $3.59\times 10^{-2}$ $\pm 1.55\times 10^{-2}$ &                                            &  $49.8\%$ &          &        \\
     \textbf{rgnp-lr} $\alpha~2, \beta~0$ & 512 & $2.39\times 10^{-2}$ $\pm 1.16\times 10^{-2}$ &  $4.07\times 10^{-3}$ $\pm 3.98\times 10^{-3}$ &  $50.4\%$ & $66.5\%$ & $306.2$\\
                                         &     & $3.62\times 10^{-2}$ $\pm 1.54\times 10^{-2}$ &                                            &  $58.0\%$ &          &        \\
\bottomrule
\end{tabular}
\end{sc}
\end{small}
\end{center}
\vskip -0.1in
\end{table*}
\subsection{Comparison of Physics \& Data Informed Model}
In this section, we show the results from the physics and data informed model compared to the nonlinear Poisson 1D problem. The setups are similar to the previous section but now the loss for the proposed model is given by \eqref{eq:elbo_phys_data} and we incorporate noisy observation of solution fields in the inference. We use 1k noisy sample solutions measured at 60 locations with a noise standard deviation $\sigma_n = 0.05$. In Fig~\ref{fig:exampleNoisyData} we show 5 samples of observed solution fields. The algorithm only sees the scatter points, not the solid ground-truth line. The losses for the other methods against which we compare are modified to include a data-fit term as in \citet{wang2021learning} to have a trade-off between fitting the differential operator and the observations. When incorporating data from noisy observations of solution fields, it is of crucial importance to have an inference scheme capable of correctly characterizing the uncertainty inherent in the observations~\cite{kennedy2001bayesian} to maximise the accuracy of the inference. A deterministic formulation is not able to take account of the noise in the observations and thus produces a single estimate of the parameters that might over-fit to the noise in the observations.
\begin{figure}[h!]
    \centering
    \vskip 0.2in
    \subfigure[Streamlines FE]{
        \includegraphics[width=0.48\columnwidth]{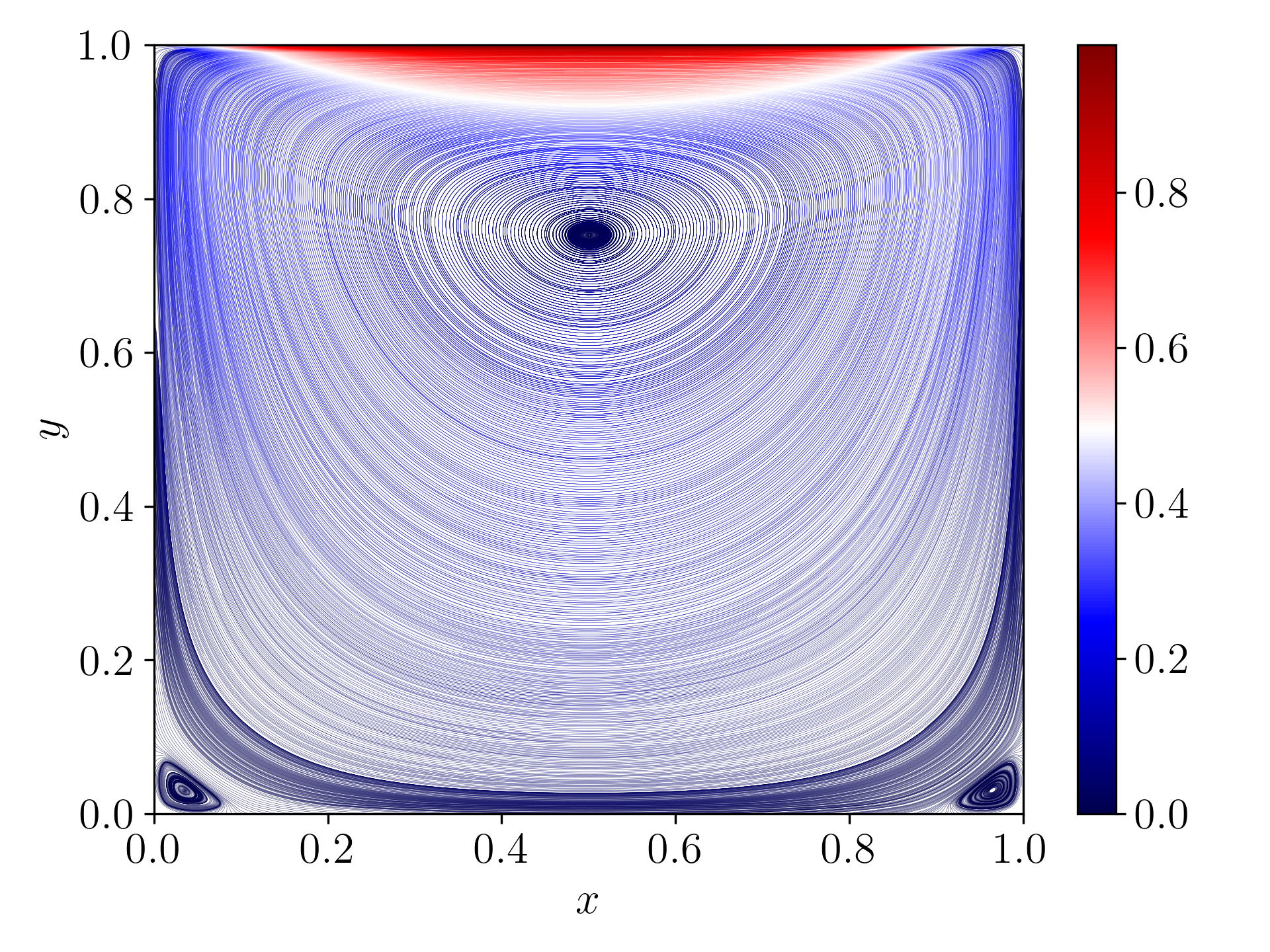}}  
    \subfigure[Streamlines $\alpha$-Net diag.]{
        \includegraphics[width=0.48\columnwidth]{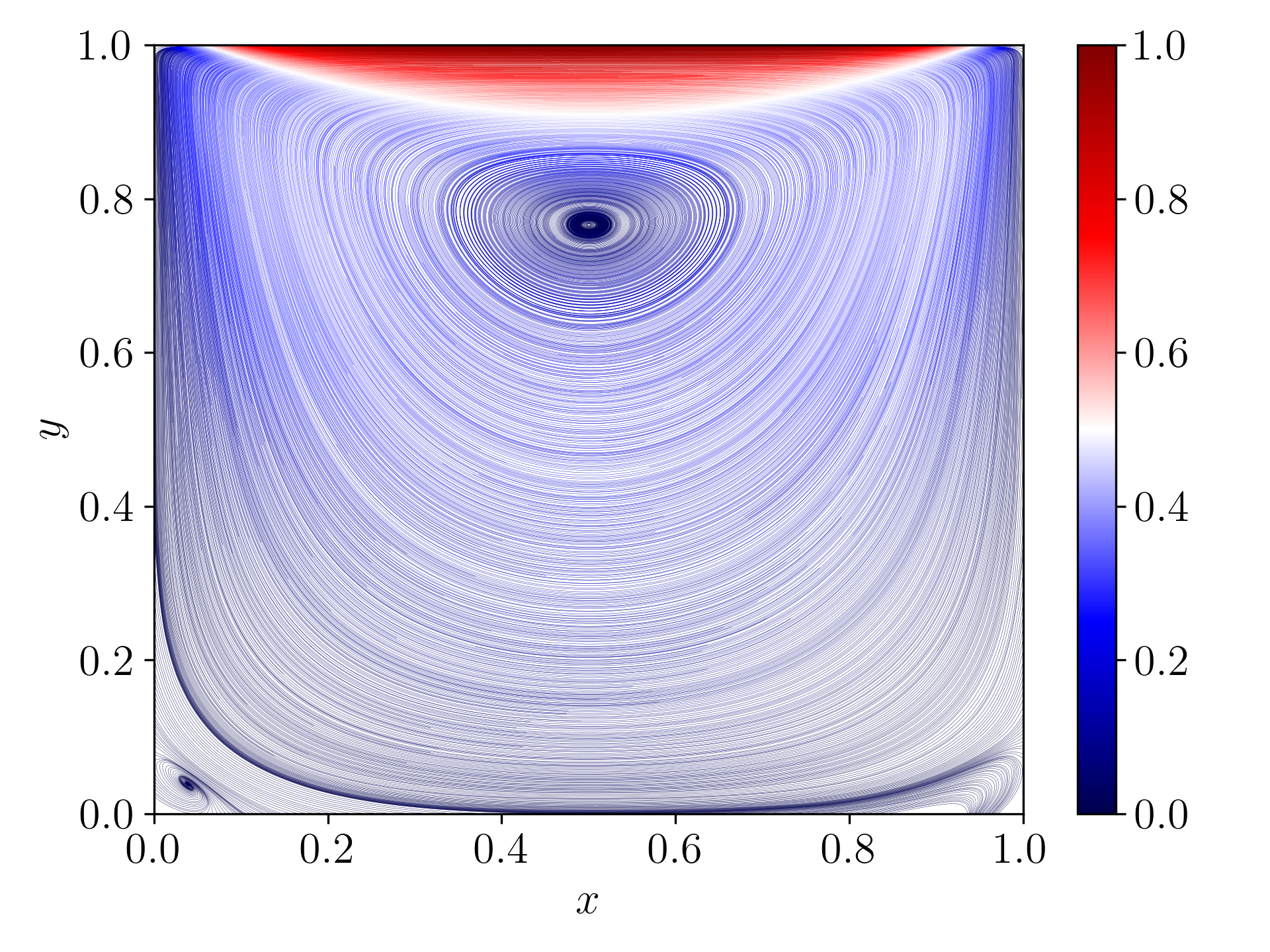}}  
    \\
    \subfigure[$u_1$ FE]{
        \includegraphics[width=0.48\columnwidth]{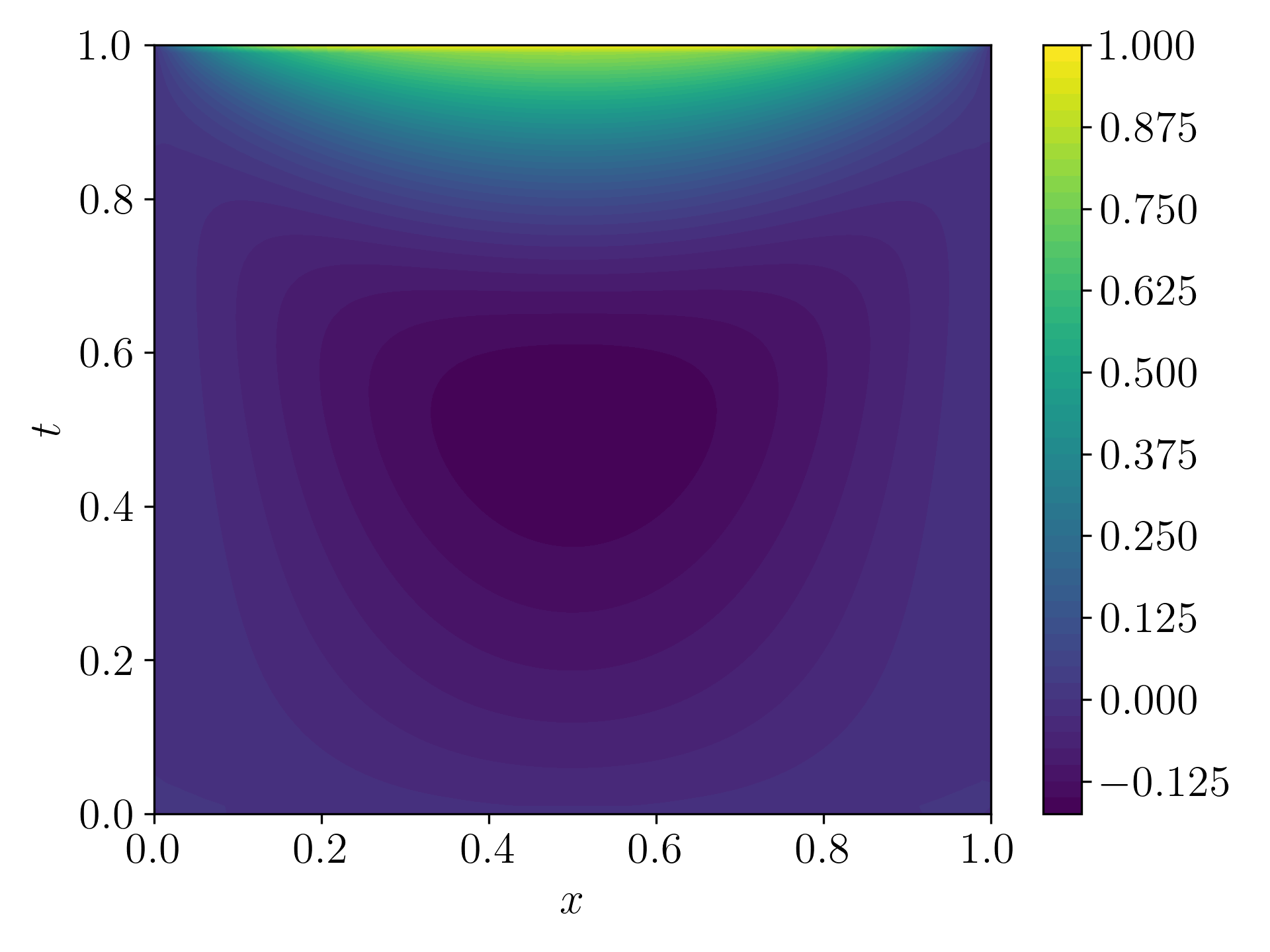}}   
    \subfigure[$u_1$ $\alpha$-Net diag.]{
        \includegraphics[width=0.48\columnwidth]{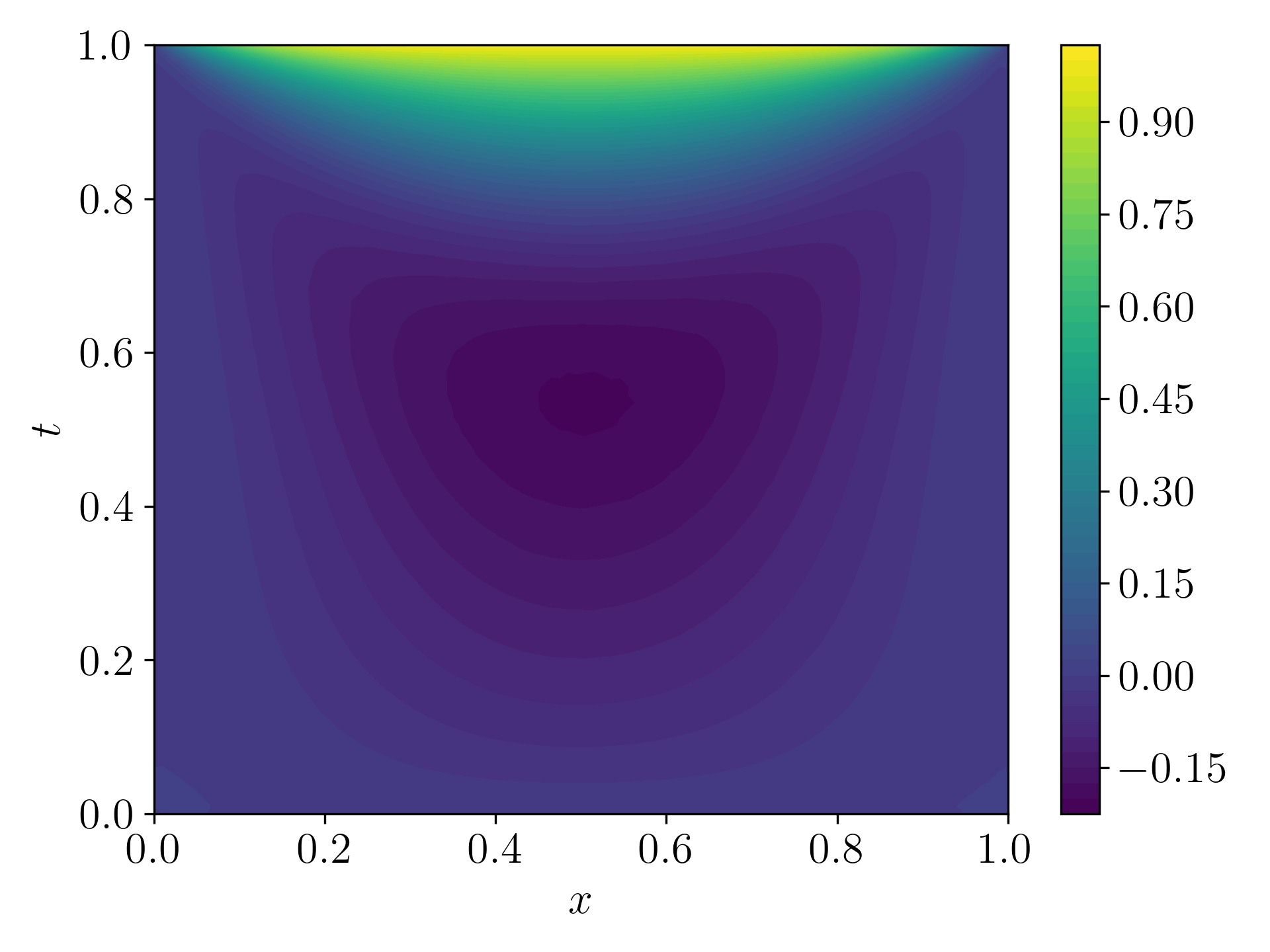}}
        \\
        \subfigure[$u_2$ FEM]{
        \includegraphics[width=0.48\columnwidth]{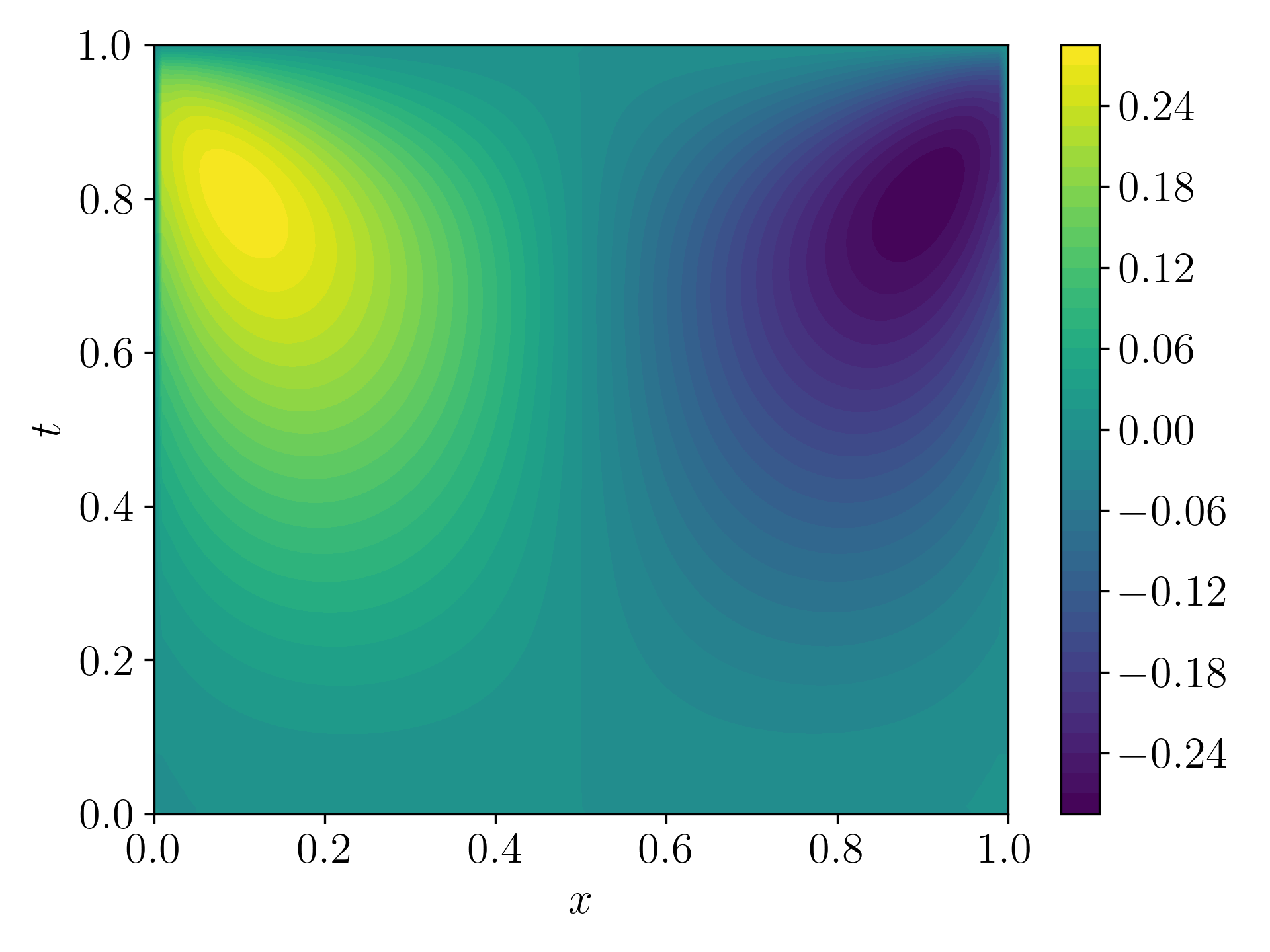}}    
    \subfigure[$u_2$ $\alpha$-Net diag.]{
        \includegraphics[width=0.48\columnwidth]{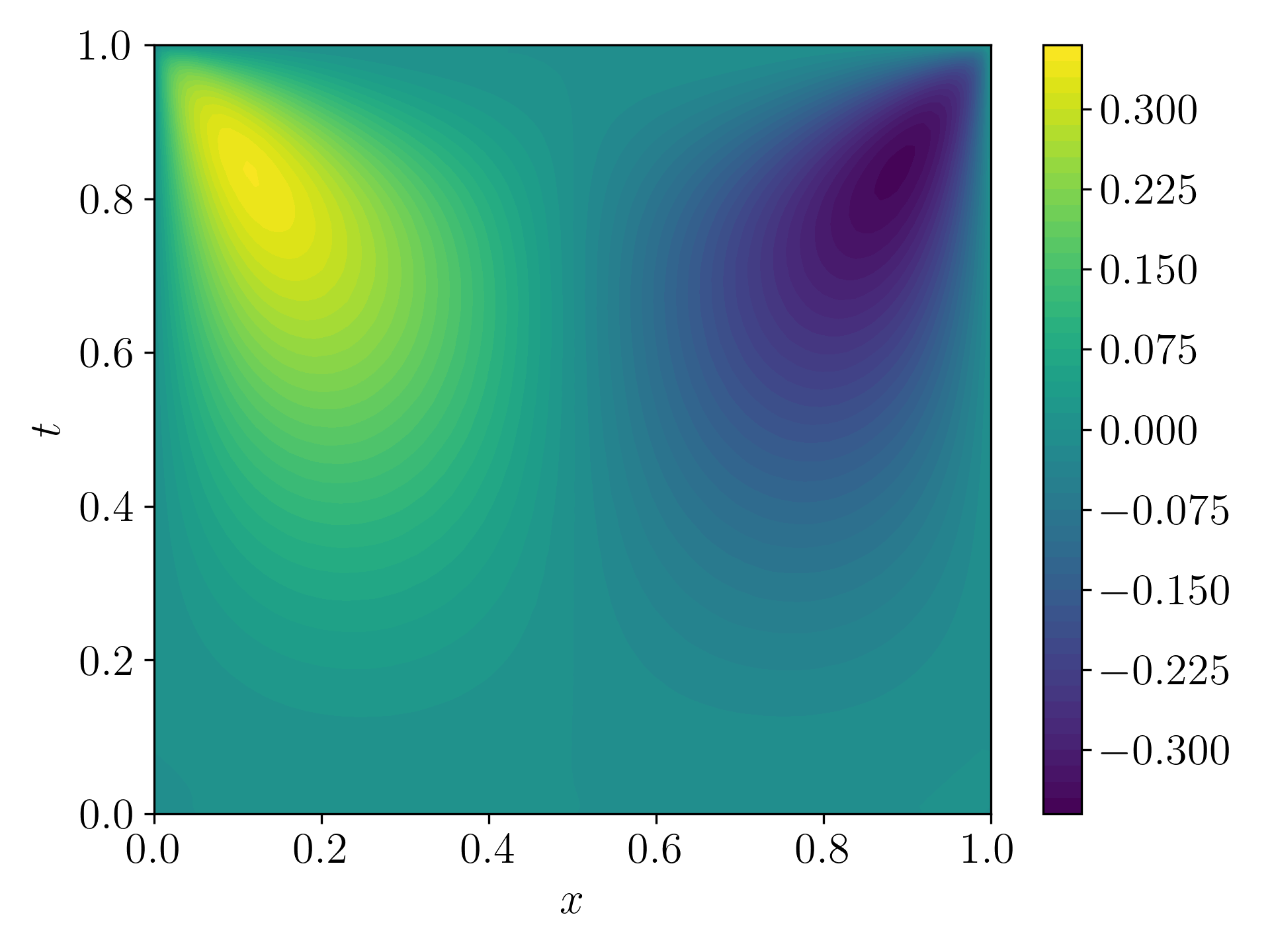}} 
    \caption{A comparison of the mean solution fields of the $\alpha$-Net with a diagonal covariance and the FE solution. }
    \vskip -0.2in
\end{figure}

\subsection{Navier-Stokes Lid Driven Cavity Flow}\label{sec:nslid_experiment}
We apply our method to a challenging parametric Navier-Stokes problem. We define the parametric differential operator and boundary conditions for a time-dependent lid-driven cavity flow example using the incompressible Navier-Stokes equations. Here the parameters correspond to the dynamic viscosity and the fluid density.
In Table~\ref{tbl:comparisonsPhysOnly} we compute these quantities for 100 independent samples of $z$ and $w$ drawn from the priors. All models are run for 100k gradient update steps. We note that obtaining reliable and accurate reference solutions to Navier-Stokes problems is very challenging.

\section{Conclusion}
In this paper, we propose a new framework of random collocation neural processes for solving forward and inverse parametric physics problems. Our method leverages spatial statistics, variational inference, neural processes, and a proposed grid-invariant convolutional network to solve forward and inverse problems in a probabilistically coherent manner. Our method is physics informed at training time and can incorporate noisy observations of solutions fields from arbitrary grids in a statistically principled way. We test our method on a nonlinear Poisson problem, the Burgers equation, and the incompressible Navier-Stokes equations. We further compare our method with a series of alternative physics and data-informed methods. We find our method is highly competitive with other approaches in terms of accuracy, uncertainty quantification, and compute time. This strongly supports the probabilistic treatment of collocation grids for the physics-informed solution of parametric PDEs.

The uncertainty captured by the proposed methodology reflects the confidence of the model with respect to the given solution fields and parameters. When using these models in practice, the application expert can then use this uncertainty to gauge the reliability of the predictions from the model, an important feature due to the black-box nature of deep learning models. In essence, a practitioner has information helping them assess their confidence in the accuracy of the given solution fields and whether more training is required or if they should use a different solution method outright. Furthermore, the inverse problems are often ill-posed and a range of parameters may yield the observed solution fields; a uncertainty quantification framework can capture this while a deterministic approach cannot.

The performance of our approach on the Navier-Stokes examples points to some limitations in using uniform distributions over domains for sampling the collocation points. The distributions $p(\bX)$ can be adapted to sample more points close to boundaries to better capture complex boundary effects; this can be implemented within our framework. Divergence enforcing \cite{richterneural} methods could potentially also be used to accelerate convergence. Chebyshev network architectures could also be investigated \cite{tang2023physics} as an alternative architecture for faster and more regularized residual computation. Further possible extensions include the incorporation of CAN-PINN collocation methods \cite{chiu2022can}. The GP formulation of our method implies that other GP methods such as deep kernels \cite{wilson2016deep} and sparse GPs could be leveraged \cite{snelson2005sparse, titsias2009variational}. We could also make use of variational weak forms in the residual \cite{kharazmi2019variational} to lower the differentiability order of the PDEs and potentially increase learning stability. The generality of the proposed framework implies that it can be readily extended to incorporate many of the newest advances in physics informed machine learning. 

\section*{Acknowledgements}

A. V. was supported by the Baxter \& Alma Ricard Foundation Scholarship. I. K. was funded by a Biometrika Fellowship awarded by the Biometrika Trust. Y. P. was supported by a Roth Scholarship funded by the Department of Mathematics, Imperial College London. F. C. was supported by Wave 1 of The UKRI Strategic Priorities Fund under the EPSRC Grant EP/T001569/1, particularly the “Digital twins for complex engineering systems" theme within that grant, and The Alan Turing Institute. M. G was supported by a Royal Academy of Engineering Research Chair, and EPSRC grants EP/W005816/1, EP/V056441/1, EP/V056522/1, EP/T000414/1, EP/R018413/2, EP/R034710/1, EP/R004889/1. This work has been supported by The Alan Turing Institute through the Theory and Methods Challenge Fortnights event ``Accelerating generative models and nonconvex optimisation'', which took place on 6-10 June 2022 and 5-9 Sep 2022 at The Alan Turing Institute headquarters. We thank the reviewers for their insightful comments.




\bibliography{bibli}

\begin{thebibliography}{64}
\providecommand{\natexlab}[1]{#1}
\providecommand{\url}[1]{\texttt{#1}}
\expandafter\ifx\csname urlstyle\endcsname\relax
  \providecommand{\doi}[1]{doi: #1}\else
  \providecommand{\doi}{doi: \begingroup \urlstyle{rm}\Url}\fi

\bibitem[Ardizzone et~al.(2018)Ardizzone, Kruse, Wirkert, Rahner, Pellegrini,
  Klessen, Maier-Hein, Rother, and K{\"o}the]{ardizzone2018analyzing}
Ardizzone, L., Kruse, J., Wirkert, S., Rahner, D., Pellegrini, E.~W., Klessen,
  R.~S., Maier-Hein, L., Rother, C., and K{\"o}the, U.
\newblock Analyzing inverse problems with invertible neural networks.
\newblock \emph{arXiv preprint arXiv:1808.04730}, 2018.

\bibitem[Belov(2012)]{belov2012inverse}
Belov, Y.~Y.
\newblock Inverse problems for partial differential equations.
\newblock In \emph{Inverse Problems for Partial Differential Equations}. De
  Gruyter, 2012.

\bibitem[Bhattacharya et~al.(2021)Bhattacharya, Hosseini, Kovachki, and
  Stuart]{bhattacharya2020model}
Bhattacharya, K., Hosseini, B., Kovachki, N.~B., and Stuart, A.~M.
\newblock Model {Reduction} {And} {Neural} {Networks} {For} {Parametric}
  {PDEs}.
\newblock \emph{The SMAI journal of computational mathematics}, 7, 2021.
\newblock \doi{10.5802/smai-jcm.74}.

\bibitem[Botella \& Peyret(1998)Botella and Peyret]{botella1998benchmark}
Botella, O. and Peyret, R.
\newblock Benchmark spectral results on the lid-driven cavity flow.
\newblock \emph{Computers \& Fluids}, 27\penalty0 (4):\penalty0 421--433, 1998.

\bibitem[Brunton et~al.(2016)Brunton, Proctor, and
  Kutz]{brunton2016discovering}
Brunton, S.~L., Proctor, J.~L., and Kutz, J.~N.
\newblock Discovering governing equations from data by sparse identification of
  nonlinear dynamical systems.
\newblock \emph{Proceedings of the national academy of sciences}, 113\penalty0
  (15):\penalty0 3932--3937, 2016.

\bibitem[Burt et~al.(2020)Burt, Ober, Garriga-Alonso, and van~der
  Wilk]{burt2020understanding}
Burt, D.~R., Ober, S.~W., Garriga-Alonso, A., and van~der Wilk, M.
\newblock Understanding variational inference in function-space.
\newblock In \emph{Third Symposium on Advances in Approximate Bayesian
  Inference}, 2020.

\bibitem[Cai(2001)]{cai2001weighted}
Cai, Z.
\newblock Weighted nadaraya--watson regression estimation.
\newblock \emph{Statistics \& probability letters}, 51\penalty0 (3):\penalty0
  307--318, 2001.

\bibitem[Chen et~al.(2021)Chen, Hosseini, Owhadi, and Stuart]{chen2021solving}
Chen, Y., Hosseini, B., Owhadi, H., and Stuart, A.~M.
\newblock Solving and learning nonlinear {PDE}s with {G}aussian processes.
\newblock \emph{Journal of Computational Physics}, 447:\penalty0 110668, 2021.

\bibitem[Chiu et~al.(2022)Chiu, Wong, Ooi, Dao, and Ong]{chiu2022can}
Chiu, P.-H., Wong, J.~C., Ooi, C., Dao, M.~H., and Ong, Y.-S.
\newblock Can-pinn: A fast physics-informed neural network based on
  coupled-automatic--numerical differentiation method.
\newblock \emph{Computer Methods in Applied Mechanics and Engineering},
  395:\penalty0 114909, 2022.

\bibitem[Cressie \& Moores(2021)Cressie and Moores]{cressie2021spatial}
Cressie, N. and Moores, M.~T.
\newblock Spatial statistics.
\newblock \emph{arXiv preprint arXiv:2105.07216}, 2021.

\bibitem[Dehaene \& Brossard(2021)Dehaene and Brossard]{dehaene2021re}
Dehaene, D. and Brossard, R.
\newblock Re-parameterizing {VAE}s for stability.
\newblock \emph{arXiv preprint arXiv:2106.13739}, 2021.

\bibitem[Ding \& Zhou(2007)Ding and Zhou]{ding2007eigenvalues}
Ding, J. and Zhou, A.
\newblock Eigenvalues of rank-one updated matrices with some applications.
\newblock \emph{Applied Mathematics Letters}, 20\penalty0 (12):\penalty0
  1223--1226, 2007.

\bibitem[Duffin(2022)]{duffin2022statistical}
Duffin, C.
\newblock Statistical finite element methods for nonlinear {PDE}s.
\newblock 2022.

\bibitem[Fanaskov \& Oseledets(2022)Fanaskov and
  Oseledets]{fanaskov2022spectral}
Fanaskov, V. and Oseledets, I.
\newblock Spectral neural operators.
\newblock \emph{arXiv preprint arXiv:2205.10573}, 2022.

\bibitem[Gao et~al.(2022)Gao, Zahr, and Wang]{gao2022physics}
Gao, H., Zahr, M.~J., and Wang, J.-X.
\newblock Physics-informed graph neural {G}alerkin networks: A unified
  framework for solving {PDE}-governed forward and inverse problems.
\newblock \emph{Computer Methods in Applied Mechanics and Engineering},
  390:\penalty0 114502, 2022.

\bibitem[Garnelo et~al.(2018{\natexlab{a}})Garnelo, Rosenbaum, Maddison,
  Ramalho, Saxton, Shanahan, Teh, Rezende, and Eslami]{garnelo2018conditional}
Garnelo, M., Rosenbaum, D., Maddison, C., Ramalho, T., Saxton, D., Shanahan,
  M., Teh, Y.~W., Rezende, D., and Eslami, S.~A.
\newblock Conditional neural processes.
\newblock In \emph{International Conference on Machine Learning}, pp.\
  1704--1713. PMLR, 2018{\natexlab{a}}.

\bibitem[Garnelo et~al.(2018{\natexlab{b}})Garnelo, Schwarz, Rosenbaum, Viola,
  Rezende, Eslami, and Teh]{garnelo2018neural}
Garnelo, M., Schwarz, J., Rosenbaum, D., Viola, F., Rezende, D.~J., Eslami, S.,
  and Teh, Y.~W.
\newblock Neural processes.
\newblock \emph{arXiv preprint arXiv:1807.01622}, 2018{\natexlab{b}}.

\bibitem[Glyn-Davies et~al.(2022)Glyn-Davies, Duffin, Akyildiz, and
  Girolami]{glyn2022phi}
Glyn-Davies, A., Duffin, C., Akyildiz, {\"O}.~D., and Girolami, M.
\newblock {$\Phi$}-{DVAE}: {L}earning {P}hysically {I}nterpretable
  {R}epresentations with {N}onlinear {F}iltering.
\newblock \emph{arXiv preprint arXiv:2209.15609}, 2022.

\bibitem[Kaltenbach et~al.(2023)Kaltenbach, Perdikaris, and
  Koutsourelakis]{kaltenbach2023semi}
Kaltenbach, S., Perdikaris, P., and Koutsourelakis, P.-S.
\newblock Semi-supervised invertible neural operators for bayesian inverse
  problems.
\newblock \emph{Computational Mechanics}, pp.\  1--20, 2023.

\bibitem[Kennedy \& O'Hagan(2001)Kennedy and O'Hagan]{kennedy2001bayesian}
Kennedy, M.~C. and O'Hagan, A.
\newblock Bayesian calibration of computer models.
\newblock \emph{Journal of the Royal Statistical Society: Series B (Statistical
  Methodology)}, 63\penalty0 (3):\penalty0 425--464, 2001.

\bibitem[Kharazmi et~al.(2019)Kharazmi, Zhang, and
  Karniadakis]{kharazmi2019variational}
Kharazmi, E., Zhang, Z., and Karniadakis, G.~E.
\newblock Variational physics-informed neural networks for solving partial
  differential equations.
\newblock \emph{arXiv preprint arXiv:1912.00873}, 2019.

\bibitem[Kingma et~al.(2015)Kingma, Ba, Bengio, and LeCun]{kingma20153rd}
Kingma, D.~P., Ba, J., Bengio, Y., and LeCun, Y.
\newblock 3rd international conference on learning representations.
\newblock \emph{ICLR, San Diego}, 2015.

\bibitem[Li et~al.(2020)Li, Kovachki, Azizzadenesheli, Liu, Bhattacharya,
  Stuart, and Anandkumar]{li2020fourier}
Li, Z., Kovachki, N., Azizzadenesheli, K., Liu, B., Bhattacharya, K., Stuart,
  A., and Anandkumar, A.
\newblock Fourier neural operator for parametric partial differential
  equations.
\newblock \emph{arXiv preprint arXiv:2010.08895}, 2020.

\bibitem[Li et~al.(2021)Li, Zheng, Kovachki, Jin, Chen, Liu, Azizzadenesheli,
  and Anandkumar]{li2021physics}
Li, Z., Zheng, H., Kovachki, N., Jin, D., Chen, H., Liu, B., Azizzadenesheli,
  K., and Anandkumar, A.
\newblock Physics-informed neural operator for learning partial differential
  equations.
\newblock \emph{arXiv preprint arXiv:2111.03794}, 2021.

\bibitem[Logg et~al.(2012)Logg, Mardal, and Wells]{logg2012automated}
Logg, A., Mardal, K.-A., and Wells, G.
\newblock \emph{Automated solution of differential equations by the finite
  element method: The FEniCS book}, volume~84.
\newblock Springer Science \& Business Media, 2012.

\bibitem[Long et~al.(2022)Long, Wang, Krishnapriyan, Kirby, Zhe, and
  Mahoney]{pmlr-v162-long22a}
Long, D., Wang, Z., Krishnapriyan, A., Kirby, R., Zhe, S., and Mahoney, M.
\newblock {A}uto{IP}: A united framework to integrate physics into {G}aussian
  processes.
\newblock In Chaudhuri, K., Jegelka, S., Song, L., Szepesvari, C., Niu, G., and
  Sabato, S. (eds.), \emph{Proceedings of the 39th International Conference on
  Machine Learning}, volume 162 of \emph{Proceedings of Machine Learning
  Research}, pp.\  14210--14222. PMLR, 17--23 Jul 2022.

\bibitem[Lu et~al.(2021{\natexlab{a}})Lu, Jin, Pang, Zhang, and
  Karniadakis]{lu2021learning}
Lu, L., Jin, P., Pang, G., Zhang, Z., and Karniadakis, G.~E.
\newblock Learning nonlinear operators via deeponet based on the universal
  approximation theorem of operators.
\newblock \emph{Nature Machine Intelligence}, 3\penalty0 (3):\penalty0
  218--229, 2021{\natexlab{a}}.

\bibitem[Lu et~al.(2021{\natexlab{b}})Lu, Pestourie, Yao, Wang, Verdugo, and
  Johnson]{lu2021physics}
Lu, L., Pestourie, R., Yao, W., Wang, Z., Verdugo, F., and Johnson, S.~G.
\newblock Physics-informed neural networks with hard constraints for inverse
  design.
\newblock \emph{SIAM Journal on Scientific Computing}, 43\penalty0
  (6):\penalty0 B1105--B1132, 2021{\natexlab{b}}.

\bibitem[Lu et~al.(2022)Lu, Meng, Cai, Mao, Goswami, Zhang, and
  Karniadakis]{lu2022comprehensive}
Lu, L., Meng, X., Cai, S., Mao, Z., Goswami, S., Zhang, Z., and Karniadakis,
  G.~E.
\newblock A comprehensive and fair comparison of two neural operators (with
  practical extensions) based on fair data.
\newblock \emph{Computer Methods in Applied Mechanics and Engineering},
  393:\penalty0 114778, 2022.

\bibitem[Markou et~al.(2022)Markou, Requeima, Bruinsma, Vaughan, and
  Turner]{markou2022practical}
Markou, S., Requeima, J., Bruinsma, W., Vaughan, A., and Turner, R.~E.
\newblock Practical conditional neural process via tractable dependent
  predictions.
\newblock In \emph{International Conference on Learning Representations}, 2022.

\bibitem[Pang et~al.(2019)Pang, Yang, and Karniadakis]{pang2019neural}
Pang, G., Yang, L., and Karniadakis, G.~E.
\newblock Neural-net-induced {G}aussian process regression for function
  approximation and {PDE} solution.
\newblock \emph{Journal of Computational Physics}, 384:\penalty0 270--288,
  2019.

\bibitem[Petersen et~al.(2008)Petersen, Pedersen, et~al.]{petersen2008matrix}
Petersen, K.~B., Pedersen, M.~S., et~al.
\newblock The matrix cookbook.
\newblock \emph{Technical University of Denmark}, 7\penalty0 (15):\penalty0
  510, 2008.

\bibitem[Quarteroni \& Valli(2008)Quarteroni and
  Valli]{quarteroni2008numerical}
Quarteroni, A. and Valli, A.
\newblock \emph{Numerical approximation of partial differential equations}.
\newblock Springer, 2008.

\bibitem[Rahman et~al.(2022)Rahman, Johnson, and Rao]{rahman2022using}
Rahman, S., Johnson, V.~E., and Rao, S.~S.
\newblock Using the left gram matrix to cluster high dimensional data.
\newblock \emph{arXiv preprint arXiv:2202.08236}, 2022.

\bibitem[Raissi \& Karniadakis(2018)Raissi and Karniadakis]{raissi2018hidden}
Raissi, M. and Karniadakis, G.~E.
\newblock Hidden physics models: Machine learning of nonlinear partial
  differential equations.
\newblock \emph{Journal of Computational Physics}, 357:\penalty0 125--141,
  2018.

\bibitem[Raissi et~al.(2019)Raissi, Perdikaris, and
  Karniadakis]{raissi2019physics}
Raissi, M., Perdikaris, P., and Karniadakis, G.~E.
\newblock Physics-informed neural networks: A deep learning framework for
  solving forward and inverse problems involving nonlinear partial differential
  equations.
\newblock \emph{Journal of Computational physics}, 378:\penalty0 686--707,
  2019.

\bibitem[Ramachandran et~al.(2017)Ramachandran, Zoph, and
  Le]{ramachandran2017searching}
Ramachandran, P., Zoph, B., and Le, Q.~V.
\newblock Searching for activation functions.
\newblock \emph{arXiv preprint arXiv:1710.05941}, 2017.

\bibitem[Rao et~al.(2021)Rao, Sun, and Liu]{rao2021physics}
Rao, C., Sun, H., and Liu, Y.
\newblock Physics-informed deep learning for computational elastodynamics
  without labeled data.
\newblock \emph{Journal of Engineering Mechanics}, 147\penalty0 (8):\penalty0
  04021043, 2021.

\bibitem[Richter-Powell et~al.(2022)Richter-Powell, Lipman, and
  Chen]{richterneural}
Richter-Powell, J., Lipman, Y., and Chen, R.~T.
\newblock Neural conservation laws: A divergence-free perspective.
\newblock In \emph{Advances in Neural Information Processing Systems}, 2022.

\bibitem[Ripley(2005)]{ripley2005spatial}
Ripley, B.~D.
\newblock \emph{Spatial statistics}.
\newblock John Wiley \& Sons, 2005.

\bibitem[Rixner \& Koutsourelakis(2021)Rixner and
  Koutsourelakis]{rixner2021probabilistic}
Rixner, M. and Koutsourelakis, P.-S.
\newblock A probabilistic generative model for semi-supervised training of
  coarse-grained surrogates and enforcing physical constraints through virtual
  observables.
\newblock \emph{Journal of Computational Physics}, 434:\penalty0 110218, 2021.

\bibitem[Rudner et~al.(2021)Rudner, Chen, Teh, and Gal]{rudner2021tractable}
Rudner, T.~G., Chen, Z., Teh, Y.~W., and Gal, Y.
\newblock Tractable function-space variational inference in bayesian neural
  networks.
\newblock In \emph{Advances in Neural Information Processing Systems}, 2021.

\bibitem[Scholkopf et~al.(1999)Scholkopf, Mika, Burges, Knirsch, Muller,
  Ratsch, and Smola]{scholkopf1999input}
Scholkopf, B., Mika, S., Burges, C.~J., Knirsch, P., Muller, K.-R., Ratsch, G.,
  and Smola, A.~J.
\newblock Input space versus feature space in kernel-based methods.
\newblock \emph{IEEE transactions on neural networks}, 10\penalty0
  (5):\penalty0 1000--1017, 1999.

\bibitem[Snelson \& Ghahramani(2005)Snelson and Ghahramani]{snelson2005sparse}
Snelson, E. and Ghahramani, Z.
\newblock Sparse {G}aussian processes using pseudo-inputs.
\newblock \emph{Advances in neural information processing systems}, 18, 2005.

\bibitem[Stuart(2010)]{stuart_2010}
Stuart, A.~M.
\newblock Inverse problems: A {B}ayesian perspective.
\newblock \emph{Acta Numerica}, 19:\penalty0 451–559, 2010.
\newblock \doi{10.1017/S0962492910000061}.

\bibitem[Sukumar \& Srivastava(2022)Sukumar and Srivastava]{sukumar2022exact}
Sukumar, N. and Srivastava, A.
\newblock Exact imposition of boundary conditions with distance functions in
  physics-informed deep neural networks.
\newblock \emph{Computer Methods in Applied Mechanics and Engineering},
  389:\penalty0 114333, 2022.

\bibitem[Sun et~al.(2018)Sun, Zhang, Shi, and Grosse]{sun2018functional}
Sun, S., Zhang, G., Shi, J., and Grosse, R.
\newblock Functional variational bayesian neural networks.
\newblock In \emph{International Conference on Learning Representations}, 2018.

\bibitem[Tait \& Damoulas(2020)Tait and Damoulas]{tait2020variational}
Tait, D.~J. and Damoulas, T.
\newblock Variational autoencoding of {PDE} inverse problems.
\newblock \emph{arXiv preprint arXiv:2006.15641}, 2020.

\bibitem[Takeishi \& Kalousis(2021)Takeishi and Kalousis]{takeishi2021physics}
Takeishi, N. and Kalousis, A.
\newblock Physics-integrated variational autoencoders for robust and
  interpretable generative modeling.
\newblock \emph{Advances in Neural Information Processing Systems},
  34:\penalty0 14809--14821, 2021.

\bibitem[Tang et~al.(2023)Tang, Feng, Wu, and Xu]{tang2023physics}
Tang, S., Feng, X., Wu, W., and Xu, H.
\newblock Physics-informed neural networks combined with polynomial
  interpolation to solve nonlinear partial differential equations.
\newblock \emph{Computers \& Mathematics with Applications}, 132:\penalty0
  48--62, 2023.

\bibitem[Titsias(2009)]{titsias2009variational}
Titsias, M.
\newblock Variational learning of inducing variables in sparse {G}aussian
  processes.
\newblock In \emph{Artificial intelligence and statistics}, pp.\  567--574.
  PMLR, 2009.

\bibitem[Tripura \& Chakraborty(2023)Tripura and
  Chakraborty]{tripura2023wavelet}
Tripura, T. and Chakraborty, S.
\newblock Wavelet neural operator for solving parametric partial differential
  equations in computational mechanics problems.
\newblock \emph{Computer Methods in Applied Mechanics and Engineering},
  404:\penalty0 115783, 2023.

\bibitem[Tronarp et~al.(2022)Tronarp, Bosch, and Hennig]{pmlr-v162-tronarp22a}
Tronarp, F., Bosch, N., and Hennig, P.
\newblock Fenrir: Physics-enhanced regression for initial value problems.
\newblock In Chaudhuri, K., Jegelka, S., Song, L., Szepesvari, C., Niu, G., and
  Sabato, S. (eds.), \emph{Proceedings of the 39th International Conference on
  Machine Learning}, volume 162 of \emph{Proceedings of Machine Learning
  Research}, pp.\  21776--21794. PMLR, 17--23 Jul 2022.

\bibitem[Vadeboncoeur et~al.(2022)Vadeboncoeur, Akyildiz, Kazlauskaite,
  Girolami, and Cirak]{vadeboncoeur2022deep}
Vadeboncoeur, A., Akyildiz, {\"O}.~D., Kazlauskaite, I., Girolami, M., and
  Cirak, F.
\newblock Deep probabilistic models for forward and inverse problems in
  parametric {PDE}s.
\newblock \emph{arXiv preprint arXiv:2208.04856}, 2022.

\bibitem[Wang et~al.(2021{\natexlab{a}})Wang, Wang, and
  Perdikaris]{wang2021eigenvector}
Wang, S., Wang, H., and Perdikaris, P.
\newblock On the eigenvector bias of fourier feature networks: From regression
  to solving multi-scale {PDE}s with physics-informed neural networks.
\newblock \emph{Computer Methods in Applied Mechanics and Engineering},
  384:\penalty0 113938, 2021{\natexlab{a}}.

\bibitem[Wang et~al.(2021{\natexlab{b}})Wang, Wang, and
  Perdikaris]{wang2021learning}
Wang, S., Wang, H., and Perdikaris, P.
\newblock Learning the solution operator of parametric partial differential
  equations with physics-informed {D}eep{ON}ets.
\newblock \emph{Science advances}, 7\penalty0 (40):\penalty0 eabi8605,
  2021{\natexlab{b}}.

\bibitem[Welling \& Kingma(2014)Welling and Kingma]{welling2014auto}
Welling, M. and Kingma, D.~P.
\newblock Auto-encoding variational bayes.
\newblock In \emph{ICLR}, 2014.

\bibitem[Williams \& Rasmussen(2006)Williams and
  Rasmussen]{williams2006gaussian}
Williams, C.~K. and Rasmussen, C.~E.
\newblock \emph{{G}aussian processes for machine learning}, volume~2.
\newblock MIT press Cambridge, MA, 2006.

\bibitem[Wilson et~al.(2016)Wilson, Hu, Salakhutdinov, and
  Xing]{wilson2016deep}
Wilson, A.~G., Hu, Z., Salakhutdinov, R., and Xing, E.~P.
\newblock Deep kernel learning.
\newblock In \emph{Artificial intelligence and statistics}, pp.\  370--378.
  PMLR, 2016.

\bibitem[Yang \& Perdikaris(2019{\natexlab{a}})Yang and
  Perdikaris]{yang2019adversarial}
Yang, Y. and Perdikaris, P.
\newblock Adversarial uncertainty quantification in physics-informed neural
  networks.
\newblock \emph{Journal of Computational Physics}, 394:\penalty0 136--152,
  2019{\natexlab{a}}.

\bibitem[Yang \& Perdikaris(2019{\natexlab{b}})Yang and
  Perdikaris]{yang2019conditional}
Yang, Y. and Perdikaris, P.
\newblock Conditional deep surrogate models for stochastic, high-dimensional,
  and multi-fidelity systems.
\newblock \emph{Computational Mechanics}, 64\penalty0 (2):\penalty0 417--434,
  2019{\natexlab{b}}.

\bibitem[Zhang et~al.(2022)Zhang, Zhang, and Lin]{zhang2022pagp}
Zhang, J., Zhang, S., and Lin, G.
\newblock {PAGP}: A physics-assisted {G}aussian process framework with active
  learning for forward and inverse problems of partial differential equations.
\newblock \emph{arXiv preprint arXiv:2204.02583}, 2022.

\bibitem[Zhao et~al.(2022)Zhao, Lindell, and Wetzstein]{zhao2022learning}
Zhao, Q., Lindell, D.~B., and Wetzstein, G.
\newblock Learning to solve {PDE}-constrained inverse problems with graph
  networks.
\newblock In \emph{ICML 2nd AI for Science Workshop}, 2022.

\bibitem[Zhong \& Meidani(2023)Zhong and Meidani]{zhong2022pi}
Zhong, W. and Meidani, H.
\newblock {PI-VAE}: Physics-informed variational auto-encoder for stochastic
  differential equations.
\newblock \emph{Computer Methods in Applied Mechanics and Engineering},
  403:\penalty0 115664, 2023.
\newblock ISSN 0045-7825.

\end{thebibliography}
\bibliographystyle{icml2023}

\newpage
\appendix
\onecolumn


\section{Physics \& Data Models}
In this section, we write out the full derivation for the physics and data informed models. We include an extra derivation for a model where we observe a nonlinear transformation of the solution field and noisy parameter observations. This results in an additional lower bound. We first derive the model for maximizing the marginal likelihood of the residual and observational data. We then derive the tractable model shown in the main paper and the model for nonlinear solution field observations.
\subsection{General Physics \& Data Informed Model}\label{sec:gen_derivation_data}
We begin by presenting the overall framework for maximizing the marginal likelihood of the residual and the data. We outline the observational model in the subsequent subsections as this depends on the nature of the data we are dealing with. Firstly, as shown in the paper for the physics informed model we write out a joint distribution over all variables of interest to maximize a marginal likelihood. The new marginal likelihood is for the residual and the data observation conditioned on the dataset inputs for these observations. We factorize the model in a similar fashion as the physics informed model including a data likelihood term. For brevity, we begin with the already discretized distributions. We have
\begin{align}
    p(\r, \y_D, \ubf, \z, \w, \bX|\z_D, \w_D, \bX_D) &= p(\r|\ubf, \z, \w)p(\y_D|\z_D, \w_D, \bX_D)p_\beta(\z|\ubf, \w, \bX) p(\ubf|\bX) p(\w) p(\bX).
\end{align}
We then write out the variational approximation to the joint over the latent variables as
\begin{align}
    q(\ubf, \z, \w, \bX) &= q_\alpha(\ubf|\z, \w, \bX) q(\z) p(\w) p(\bX).
\end{align}
Here $\y_D, \z_D, \w_D, \bX_D$ denotes the observations of the solution field, the associated physical parameter, the extra model parameter, and the observation locations of the $\y_D$ data respectively. We can then write out the evidence lower-bound on the marginal likelihood using Jensen's inequality
\begin{align}
    &\log p(\r = \b0, \y_D|\z_D, \w_D, \bX_D)\geq\cF(\alpha, \beta)\\
    &\cF(\alpha, \beta) = \int \log \frac{p(\r, \y_D, \ubf, \z, \w, \bX|\z_D, \w_D, \bX_D)}{q(\ubf, \z, \w, \bX)}q(\ubf, \z, \w, \bX)\,\md\ubf\,\md\z\,\md\w\,\md\bX.
\end{align}
Rewriting the integration for an expectation we obtain
\begin{align}
     &\cF(\alpha, \beta) = \Eb_{\ubf, \z,\w,\bX} \left[ \log p(\r, \y_D, \ubf, \z, \w, \bX|\z_D, \w_D, \bX_D)  - \log q(\ubf, \z, \w, \bX) \right].
\end{align}
We then factorize out the likelihood term from the expectation as this term does not depend on the integration variables,
\begin{align}
     &\cF(\alpha, \beta) = \log p(\y_D|\z_D, \w_D, \bX_D) + \Eb_{\ubf, \z,\w,\bX} \left[ \log p(\r, \ubf, \z, \w, \bX)  - \log q(\ubf, \z, \w, \bX) \right].
\end{align}
We note that the form of the likelihood term depends on the nature of the observation model.

\subsection{Physics and Noisy Data Informed Model}\label{app:derivation_data_tractable}
We now derive the full ELBO for an observation model where we observe noisy solution fields at point locations. We pose our observation model as
\begin{align}
    \y^i_D &= G(\z^i_D,\w^i_D, \bX^i_D) + \sigma_n\ebf_1.
\end{align}
In our case, we learn the observation operator $G(\cdot)$ is $q_\alpha(\ubf|\z_D^i, \w_D^i, \bX_D^i)$ which yields a mean and a covariance. We can write this out as the resulting observation model explicitly as
\begin{align}
    \y_D^i = \bmu_\alpha(\z_D^i, \w_D^i, \bX_D^i) + \bK_{\alpha}(\bX_D^i, \bX_D^i;\z_D^i, \w_D^i)^{\frac{1}{2}}\ebf_2 + \sigma_n \ebf_1,
\end{align}
where $\ebf_1, \ebf_2  \sim \NPDF(\b0, \bI)$. This corresponds to a product of normal distributions of the form
\begin{align}
    p(\y_D|\z_D, \w_D, \bX_D) = \prod_{i=0}^N \NPDF(\y^i_D; \bmu_\alpha(\bX^i, \z^i_D, \w^i_D), \bar{\bSigma}_\alpha(\bX^i, \z^i_D, \w^i_D) ),
\end{align}
with a new covariance that takes into account the iid Gaussian noise along with the covariance coming from the $\alpha$-distribution,
\begin{align}
    \bar{\bSigma}_\alpha(\bX, \z, \w ) = \bK_\alpha(\bX, \bX; \z, \w) + \sigma_n^2\bI.
\end{align}
We will then approximate the likelihood function over the entire dataset with a mini-batch version which we write out as
\begin{align}
    \log p(\y_D|\z_D, \w_D, \bX_D) \approx \frac{N}{|M|}\sum_{i\in M} \log \NPDF(\y^i_D; \bmu_\alpha(\bX^i, \z^i_D, \w^i_D), \bar{\bSigma}_\alpha(\bX^i, \z^i_D, \w^i_D) ).
\end{align}
Bringing all things together from Sec.~\ref{sec:gen_derivation_data} we obtain the mini-batched ELBO 
\begin{align}
    &\log p(\r, \y_D|\z_D,\w_D,\bX_D)\geq \cF(\alpha, \beta)\\
     &= \sum_{i}^N \log p(\y^i_D|\z^i_D, \w^i_D, \bX^i_D)+ \Eb_{\ubf, {\z}, {\w}, {\bX}} \left[ \log \frac{p(\r|\ubf, \z, \w, \bX)p_\beta(\z|\ubf, \w, \bX)p(\ubf|\bX)}{q_\alpha(\ubf|\z, \w, \bX)q(\z)}\right]\\
    &\approx \frac{N}{|M|}\sum_{i\in M} \log p(\y^i_D|\z^i_D, \w^i_D, \bX^i_D)+ \Eb_{\ubf, {\z}, {\w}, {\bX}} \left[ \log \frac{p(\r|\ubf, \z, \w, \bX)p_\beta(\z|\ubf, \w, \bX)p(\ubf|\bX)}{q_\alpha(\ubf|\z, \w, \bX)q(\z)}\right].
\end{align}
This objective is a lower-bound on the marginal likelihood of the observational data and the physics residual. This model relies on direct noisy observations of the solution field of interest along with knowing (or deterministically estimating) the generating parameters of the PDE yielding the observations.

\subsection{ Physics and Indirect Noisy Observations and
Noisy Parameters Informed Model}\label{app:derivation_data_intractable}
In this section we derive a model that is not included in the main paper that deals with indirect observations of the solution field and noisy measurements of the source parameters of the PDE. In this case, we deal with a general setting where we indirectly observe the solution field through some other known nonlinear mapping. One of the more salient examples of such a scenario is the noisy measurement of drag coefficients given from a fluid flow. Here the PDE describes the fluid velocity, and from a velocity field, we can compute the drag coefficient. Drag coefficients can also be easier to measure than direct solution fields.

For such a model where the parameters are also noisily measured (or estimated probabilistically through a Bayesian inverse problem), we write out the full observation model as
\begin{align}
    \y^i_D &= g\left(G(\z^i_D, \w^i_D, \bX^i_D)\right) + \ebf_{\y_D^i},\quad \ebf_{\y_D^{i}}\sim\NPDF(\b0, \bepsilon_{\y_D^{i}}^2\bI),\\
    \z_D^i   &= \tilde{\z}  + \ebf_{\z^i}, \quad \ebf_{\z^i}\sim \NPDF(\b0, \bepsilon_{\z^i}^2\bI ),\\
    \w_D^i   &= \tilde{\w}  + \ebf_{\w^i}, \quad \ebf_{\w^i}\sim \NPDF(\b0, \bepsilon_{\w^i}^2\bI ),\\
    \bX_D^i  &= \tilde{\bX} + \ebf_{\bX^i}, \quad \ebf_{\bX^i}\sim \NPDF(\b0, \bepsilon_{\bX^i}^2\bI ).
\end{align}
where $ G(\cdot)$ is $q_\alpha(\ubf| \tilde{\z},\tilde{\w}, \tilde{\bX})$.
This yields a collection of Gaussian distributions which we write out as
\begin{align}
    p(\y_D^i|\tilde{\ubf}) &= \NPDF(\y_D^i| g(\tilde{\ubf}), \bepsilon_{\y_D^i}^2\bI),\\
    p(\tilde{\z}|\z_D^i) &= \NPDF(\tilde{\z}| \z_D^i, \bepsilon_{\z_D^i}^2\bI),\\
    p(\tilde{\w}|\w_D^i) &= \NPDF(\tilde{\w}| \z_D^i, \bepsilon_{\w_D^i}^2\bI),\\
    p(\tilde{\bX}|\bX_D^i) &= \NPDF(\tilde{\bX}| \bX_D^i, \bepsilon_{\bX_D^i}^2\bI).
\end{align}
The $\tilde{\ubf}, \tilde{\z}, \tilde{\w}, \tilde{\bX}$ variables are not directly observed.
We can then write out the likelihood term of the observation data as
\begin{align}
    p(\y_D|\z_D, \w_D, \bX_D) &= \prod_i^N p(\y_D^i|\z_D^i, \w_D^i, \bX_D^i),\\
    \log p(\y_D|\z_D, \w_D, \bX_D) &= \sum_i^N \log p(\y_D^i|\z_D^i, \w_D^i, \bX_D^i),\\
    \log p(\y_D|\z_D, \w_D, \bX_D) &\approx \frac{N}{|M|}\sum_{i\in M} \log p(\y_D^i|\z_D^i, \w_D^i, \bX_D^i).
\end{align}
We then marginalize out the data variables to obtain the marginal likelihood of the data in terms of the available distributions,
\begin{align}
    \log p(\y_D^i|\z_D^i, \w_D^i, \bX_D^i) &= \log \int p(\y_D^i, \tilde{\ubf}, \tilde{\z},\tilde{\w}, \tilde{\bX}|\z_D^i, \w_D^i, \bX^i )\,\md\tilde{\ubf}\,\md\tilde{\z}\,\md\tilde{\w}\,\md\tilde{\bX}, \\
    \log p(\y_D^i|\z_D^i, \w_D^i, \bX_D^i) &= \log \int p(\y_D^i|\tilde{\ubf})q_\alpha(\tilde{\ubf}|\tilde{\z}, \tilde{\w}, \tilde{\bX})p(\tilde{\z}|\z_D^i)p(\tilde{\w}|\w_D^i)p(\tilde{\bX}|\bX_D^i)\,\md\tilde{\ubf}\,\md\tilde{\z}\,\md\tilde{\w}\,\md\tilde{\bX}.
\end{align}
We then use Jensen's inequality to obtain a lower bound on this marginal likelihood in terms of log distributions for computational convenience and apply the same mini-batching as before,
\begin{align}
    \log p(\y_D^i|\z_D^i, \w_D^i, \bX_D^i) &\geq  \Eb_{\tilde{\ubf}, \tilde{\z}, \tilde{\w}, \tilde{\bX}} \left[ \log p(\y^i_D| \tilde{\ubf}) \right],\\
    \log p(\y_D|\z_D, \w_D, \bX_D) &\geq  \sum_i^N \Eb_{\tilde{\ubf}, \tilde{\z}, \tilde{\w}, \tilde{\bX}} \left[ \log p(\y_D^i| \tilde{\ubf})\right],\\
    &\approx \frac{N}{|M|}\sum_{i\in M} \Eb_{\tilde{\ubf}, \tilde{\z}, \tilde{\w}, \tilde{\bX}}  \left[  \log p(\y_D^i| \tilde{\ubf})\right].
\end{align}
We can then use the previously derived result from Sec.~\ref{sec:gen_derivation_data} to obtain the complete ELBO on the marginal likelihood of the data and the residual,
\begin{align}
    &\log p(\r, \y_D|\z_D,\w_D,\bX_D)\geq \cF(\alpha, \beta)\\
    &= \sum_{i}^N \Eb_{\tilde{\ubf}, \tilde{\z}, \tilde{\w}, \tilde{\bX}}  \left[  \log p(\y_D^i| \tilde{\ubf})\right] + \Eb_{\ubf, {\z}, {\w}, {\bX}} \left[ \log \frac{p(\r|\ubf, \z, \w, \bX)p_\beta(\z|\ubf, \w, \bX)p(\ubf|\bX)}{q_\alpha(\ubf|\z, \w, \bX)q(\z)}\right],\\
    &\approx \frac{N}{|M|}\sum_{i\in M} \Eb_{\tilde{\ubf}, \tilde{\z}, \tilde{\w}, \tilde{\bX}}  \left[  \log p(\y_D^i| \tilde{\ubf})\right] + \Eb_{\ubf, {\z}, {\w}, {\bX}} \left[ \log \frac{p(\r|\ubf, \z, \w, \bX)p_\beta(\z|\ubf, \w, \bX)p(\ubf|\bX)}{q_\alpha(\ubf|\z, \w, \bX)q(\z)}\right].
\end{align}
With this model, the method can be generalized to incorporate data from multiple sources that are not just direct observations of the solution field. This also opens the doors to the possibility of learning missing dynamics not described in the PDEs in an experimentally feasible way.

\section{Low-Rank Covariance Matrix}\label{app:low_rank_covriance}
In this section, we discuss in detail how we can use a low-rank covariance matrix. The low-rank kernel is expressed as 
\begin{align}
    k(x, x') &= \lambda(x)\delta_{x,x'} + \langle V(x), V(x') \rangle, \label{eq:kernelLR_app}
\end{align}
where $V(x)\in \Rb^{l}$ and $l$ denotes the column dimension of the rectangular matrices resulting from this kernel. Here we show that the left Gram matrix $\bV\bV^\top$ results from this kernel
\begin{align}
     (\bV\bV^{\top})_{ij} &= \sum_l \bV_{il}\bV_{jl}, \\
    & = \sum_l V(x_i)_l V(x_j)_l, \\
    & = \langle V(x_i), V(x_j) \rangle.
\end{align}
Expanding out the definition of the kernel for a collection $\bX, \bX'$ of points we obtain
\begin{align}\label{eq:lr_covariance_matrix}
    K(\bX, \bX)  &= \bLambda + \bV\bV^{\top},\\
    \bLambda &= \text{diag}( \blambda ), \\
    \bV &\in \Rb^{n \times l}, 
\end{align}
where $ \blambda \in \Rb^{n}$ and $\bV\in\Rb^{n\times l}$. 
We now show how we can sample and evaluate the log-density efficiently without ever constructing the full $n\times n$ matrix. 

\subsection{Sampling the Low-Rank Covariance}
Sampling from a typical dense covariance matrix requires a Cholesky decomposition of the Matrix, an $O(n^3)$ operation. However, with the special structure of the low-rank kernel in \eqref{eq:lr_covariance_matrix} we can sample efficiently with
\begin{align}
    \bepsilon &= \blambda^{1/2}\odot\bepsilon_{\lambda} + \bV\bepsilon_V, \\
    \bepsilon_{\lambda}& \sim \NPDF(\b0, \bI_n) \in \Rb^{n}, \\
    \bepsilon_{V}& \sim \NPDF(\b0, \bI_l) \in \Rb^{l}.
\end{align}

\subsection{Evaluating the Log-Density}
Evaluating the log-density requires two main operations that are computationally expensive: solving a linear system of equations, and evaluating a log determinant. Usually, both of these two operations can be computed with a Cholesky decomposition which is $O(n^3)$. We show how we can evaluate the log density more efficiently by using the form of \eqref{eq:lr_covariance_matrix}. Firstly we write out the log density of an MVN, 
\begin{align}
    \log \NPDF(\bmu, \bSigma) = -\frac{n}{2}(2\pi) - \frac{1}{2}\log(|K(\bX, \bX)|) -\frac{1}{2}(\mathbf{x}-\bmu)^{\top}K(\bX, \bX)^{-1}(\mathbf{x}-\bmu).
\end{align}
We first look at how to efficiently solve the linear system by making use of the Woodbury \cite{petersen2008matrix} identity,
\begin{align}
K(\bX, \bX)^{-1}\ba = \bLambda^{-1}\ba - \bLambda^{-1}\bV(\mathbf{I}_{l}+\bV^{\top}\bLambda^{-1}\bV)^{-1}\bV^{\top}\bLambda^{-1}\ba,
\end{align}
where $\ba = \mathbf{x}-\bmu$. To compute this efficiently we make use of 
\begin{align}
    (\bLambda^{-1}\cdot\ba)_{i} &= \blambda^{-1}_i\ba_i, \\
    (\bLambda^{-1}\cdot\bV)_{ij} &= \blambda^{-1}_i\bV_{ij}.
\end{align}
The largest matrix is $l\times n$ and requires inverting a $l\times l$ matrix where $l\ll n$. We then look at how to efficiently compute the determinant of the low-rank covariance matrix. Using the matrix determinant lemma \cite{ding2007eigenvalues} we have
\begin{align}
    |K(\bX, \bX)| = \det(\bLambda + \bV\bV^{\top}) = \det(\bI_l+\bV^{\top}\bLambda^{-1}\bV)\prod_i\blambda_i.
\end{align}
This only requires taking the determinant of a $l\times l$ matrix. This can be done by computing the Cholesky of the inner matrix as,
\begin{align}
    L = \text{chol}(\bI_l + \bV^\top\bLambda^{-1}\bV),
\end{align}
and computing the log determinant as,
\begin{align}
    \log|\bSigma| = 2\text{Tr}(\log(L)).
\end{align}
To form the covariance matrix in \eqref{eq:lr_covariance_matrix} from the neural network output of the $\alpha$-Net we have 
\begin{align}
    N_\alpha(\z, \w, \bX_{1:n}) = \Obf_{1:n}, \quad \obf_i \in \Rb^{2+l}, 
\end{align}
where $N_\alpha(\cdot)$ is a the $\alpha$-Net and
\begin{align}
    \Obf_{1:n, 1} &= \bmu_{\alpha}\in \Rb^{n}, \\
    \bsigma_T(\Obf_{1:n, 2}) &= \blambda_{\alpha} \in \Rb^{n} , \\
    \Obf_{1:n, 3:3+l} &= \bV_{\alpha}\in \Rb^{n\times l} .
\end{align}
and $\bsigma_T(\cdot)$ is an exp-linear function \cite{dehaene2021re} constraining the value to be in a set positive interval typically chosen to be $[10^{-5}, 1]$. 
In Sec.~\ref{sec:nslid_experiment} we construct distributions for a vector field, in which case the distribution can be written simply as 
\begin{align}
    q_{\alpha}(\ubf^1, \hdots, \ubf^d | \z, \w, \bomega) = 
    \NPDF\left(
    \begin{bmatrix}
    \bmu^1_\alpha(\hdots)\\
    \vdots\\
    \bmu^d_\alpha(\hdots)
    \end{bmatrix}
    ;\text{diag}
    \begin{bmatrix}
    \blambda^1_\alpha(\hdots)\\
    \vdots\\
    \blambda^d_\alpha(\hdots)
    \end{bmatrix}
    +
    \begin{bmatrix}
    \bV^1_\alpha(\hdots)\\
    \vdots\\
    \bV^d_\alpha(\hdots)
    \end{bmatrix}
    \begin{bmatrix}
    \bV^1_\alpha(\hdots)\\
    \vdots\\
    \bV^d_\alpha(\hdots)
    \end{bmatrix}^\top
    \right).
\end{align}

We note that the columns of $\bV$ must be unique, else the matrix will be singular.

\section{Distribution of Normalised Squared Errors}
In this section we show boxplots of the 1000 log NSEs samples for RGNP-D (corresponding to row 1 of table 1) and DeepONet (trained with 300 collocation points, corresponding to row 7 of Table 1. - Poisson ) for the nonlinear Poisson example.

\begin{figure}[h!]
    \centering
    \includegraphics[width=0.5\textwidth]{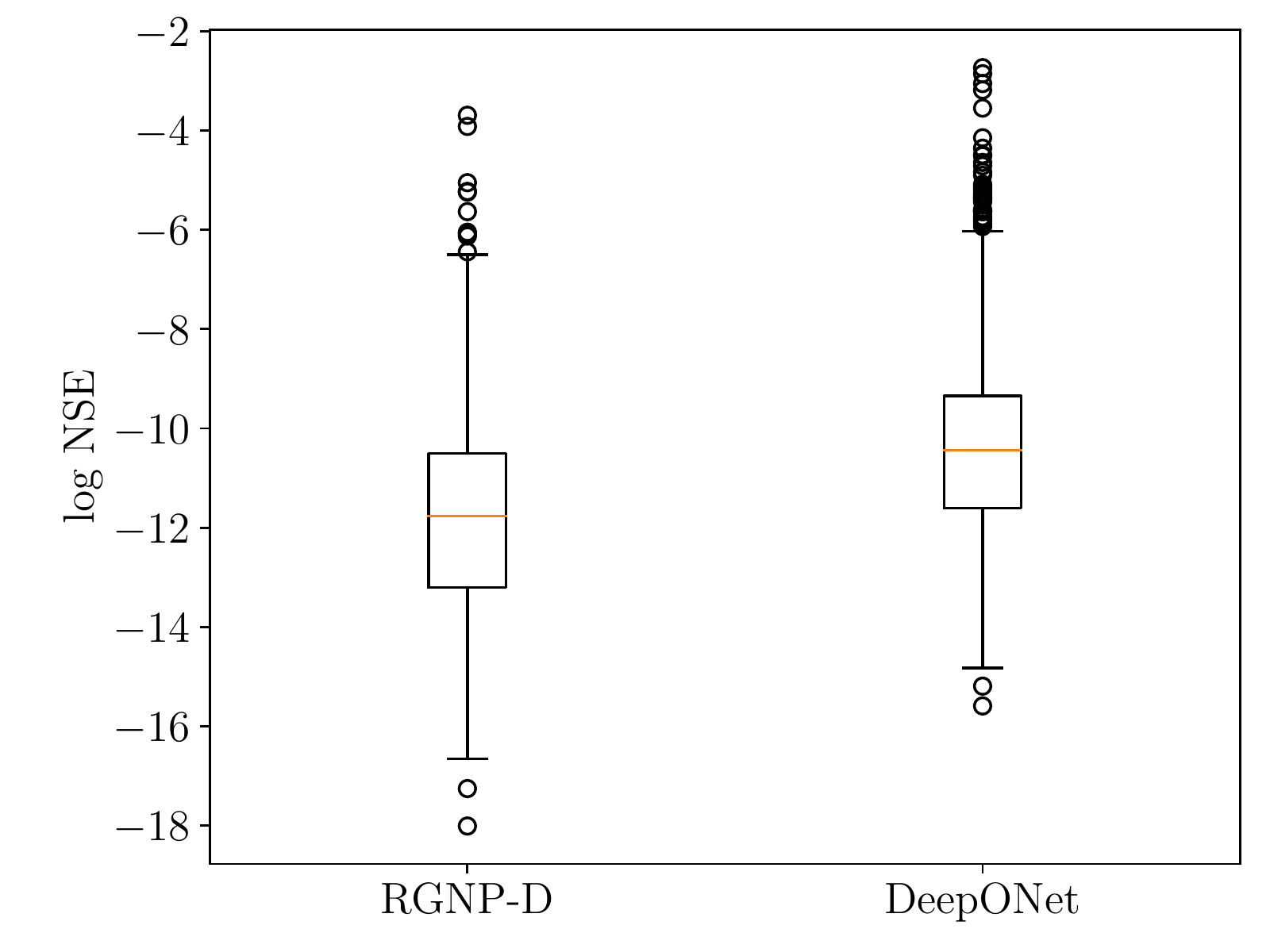}
    \caption{Box plot of log NSE for RGNP-D and DeepONet (300 N. Coll.) for the nonlinear Poisson example.}
    \label{fig:boxplot-logNSE}
\end{figure}

In Fig.~\ref{fig:boxplot-logNSE} we see the spread of the samples in the log domain. The outliers as exponentially far from the mean.




\section{Experiments}
\begin{figure}
    \vskip 0.2in
    \begin{center}
    \centerline{\includegraphics[width=0.45\textwidth, angle=270]{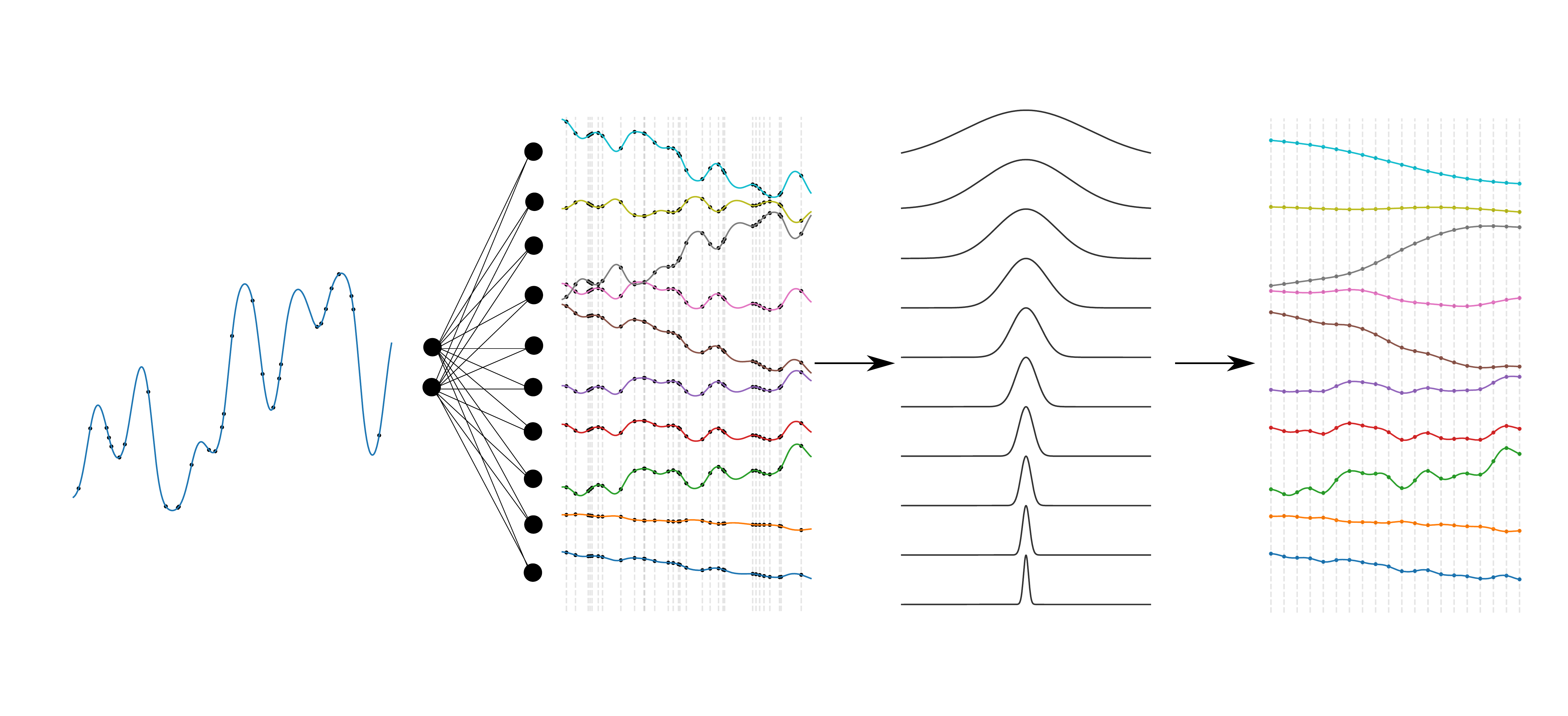}}
    \caption{A visual representation of the lifting of a signal sampled at random locations into a higher dimensional feature space where it is then interpolated with various kernels onto a fixed grid. This rich feature space evaluated at a fixed location is then passed to a convolutional neural network with a matching channel dimension.}
    \label{fig:signal_feature_space_representation}
    \end{center}
    \vskip -0.2in
\end{figure}

\begin{figure}[t]
    \vskip 0.2in
    \begin{center}
    \centerline{\includegraphics[width=0.5\textwidth]{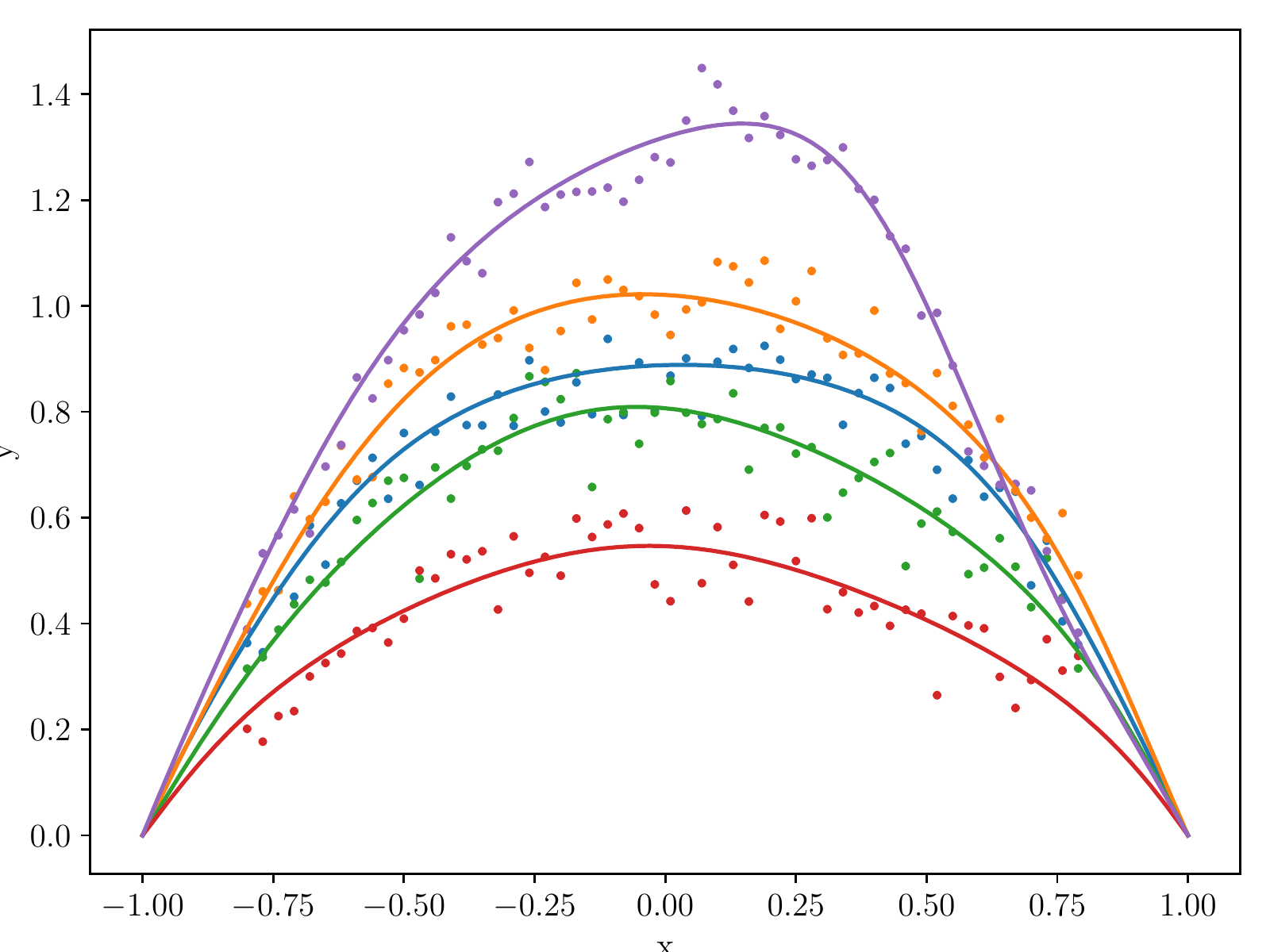}}
    \caption{Five example data samples. We show the true solution given by the solver in a solid line, and the noisy scatter data $\y_D$.}
    \label{fig:exampleNoisyData}
    \end{center}
    \vskip -0.2in
\end{figure}

The experiments were all run using TensorFlow.  All experiments were conducted with ``swish'' activation functions \cite{ramachandran2017searching}. The inversion networks are implemented with 1D, 2D, and alternated 1D \& 2D convolutional networks for the Nonlinear Poisson, Burgers, and Navier-Stokes problems, respectively. In Fig.~\ref{fig:exampleNoisyData} we show 5 example data samples. The scatter points are the data passed to the algorithm. A visual representation of the GICNet interpolation layers can be seen in Fig.~\ref{fig:signal_feature_space_representation}. 
\subsection{Nonlinear Poisson}
The $D(x)$ function used to enforce the Dirichlet boundary conditions for this problem is
\begin{align}
    D(x) = \cos\left(\frac{x\pi}{2}\right).
\end{align}
\textbf{Architecture Details:}
\begin{itemize}
    \item number of  hidden layers: 7
    \item number of neurons /  hidden layer: 300
    \item activation: swish
    \item GICNet channel dimension: 20
    \item GICNet point/dim lattice: 20
\end{itemize}
 We generate a dataset of 1000 evaluations for parameters drawn from the priors as the ground truth. It took 12.0 minutes to generate 1000 solutions using FEniCS.
\subsection{Burgers}
For the DeepONet, FNN, and GICNet we train forward net with mean absolute residual as squared mean residual is too unstable. We have 10,000 training latent parameters. For DeepONet we shorten the number of collocation evaluations for $100 \times 100$ grid as we reduce the batch size but keep the same number of iterations. Generating the 1000 sample dataset using FEniCS takes 33.5 mins. 
The $D(x, t)$ function used to enforce the Dirichlet boundary conditions for this problem is
\begin{align}
    D(x, t) = \sin\left(x\pi\right) t.
\end{align}

\textbf{Architecture Details:}
\begin{itemize}
    \item number of  hidden layers: 8
    \item number of neurons /  hidden layer: 400
    \item activation: swish
    \item GICNet channel dimension: 20
    \item GICNet point/dim lattice: 10
\end{itemize}

\subsection{Navier-Stokes}
The $D(x, t)$ function used to enforce the Dirichlet boundary conditions for this problem is
\begin{align}
    D(x,y,t) = \sin(x\pi) \sin(y\pi) t.
\end{align}
The full Navier-Stokes equations can be written more explicitly as 
\begin{align} 
    &z_0 \frac{\partial u_1}{\partial t} + z_0 \left(u_1\frac{\partial u_1}{\partial x } + u_2 \frac{\partial u_1}{\partial y }\right) + \frac{\partial p}{\partial x } - z_1 \left( \frac{\partial^2 u_1}{\partial x^2 } + \frac{\partial^2 u_1}{\partial y^2 } \right)  = 0,\nonumber \\
    &z_0 \frac{\partial u_2}{\partial t} + z_0 \left(u_1\frac{\partial u_2}{\partial x } + u_2 \frac{\partial u_2}{\partial y }\right) + \frac{\partial p}{\partial y } - z_1 \left( \frac{\partial^2 u_2}{\partial x^2 } + \frac{\partial^2 u_2}{\partial y^2 } \right)  = 0, \nonumber\\
    &\frac{\partial u_1}{\partial x} + \frac{\partial u_2}{\partial y} = 0.
\end{align}
The fully defined boundary and initial conditions are
\begin{align}
    &u_1(x,1,t) = (1-(2x-1)^6) t,\\
    &u_1(0,y,t) = u_1(1,y,t) = u_1(x,0,t) = 0,\nonumber \\
    &u_2(0,y,t) = u_2(1,y,t) = u_2(x,0,t) = u_2(x,1,t) = 0,\nonumber\\
    &u_1(x,y,0) = u_2(x,y,0) = 0,\nonumber\\
    &p(0,0,t) = 0.\nonumber
\end{align}
\textbf{Architecture Details:}
\begin{itemize}
    \item number of hidden layers: 10
    \item number of neurons / hidden layer: 200
    \item activation: swish
    \item GICNet channel dimension: 30
    \item GICNet point/dim lattice: 10
\end{itemize}

\subsection{MNSE}
The expression used for defining the MNSE used to test the methods is
\begin{align}\label{eq:mnse}
    \text{MNSE}(x, x^*) = \frac{1}{N}\sum_i^N\frac{\|x-x^*\|^2_2}{\|x^*\|^2_2},
\end{align}
where $x$ is the output of the method and $x^*$ is the ground truth.

\subsection{Hardware}
All experiments were run on an AMD Ryzen 9 5950X CPU (16 cores, 32 virtual) with 128GB memory and a Nvidia RTX 3090 (24GB VRAM) GPU. The GPU memory usage was limited to 6GB for the nonlinear Poisson problems, 10GB for the Burgers examples, and 20GB for the Navier-Stokes examples.


\end{document}